\PassOptionsToPackage{table}{xcolor}
\documentclass{article}
\usepackage{iclr2026_conference,times}

\usepackage{amsmath,amsfonts,bm}

\def\eqref#1{equation~\ref{#1}}

\def\1{\bm{1}}

\DeclareMathAlphabet{\mathsfit}{\encodingdefault}{\sfdefault}{m}{sl}
\SetMathAlphabet{\mathsfit}{bold}{\encodingdefault}{\sfdefault}{bx}{n}

\usepackage{graphicx}
\usepackage{booktabs}
\usepackage{xcolor}
\usepackage[hypertexnames=false]{hyperref}
\usepackage{url}
\usepackage{tabularx}
\usepackage{multirow}
\usepackage{makecell}
\usepackage{listings}
\usepackage{placeins}
\usepackage{float}

\hypersetup{
  colorlinks=true,
  linkcolor=blue,
  citecolor=blue,
  urlcolor=blue,
  pdftitle={LazyMem: Retrieve Broadly, Construct Selectively for Efficient Long-Term Agent Memory},
  pdfauthor={Jing Yu, Yibo Zhao, Jiaming Zhang, and Xiang Li}
}

\providecommand{\Description}[1]{}
\providecommand{\balance}{}

\lstdefinestyle{prompt}{
  basicstyle=\ttfamily\scriptsize,
  columns=fullflexible,
  keepspaces=true,
  showstringspaces=false,
  breaklines=true,
  breakatwhitespace=true,
  breakautoindent=false,
  breakindent=1em,
  morecomment=[l][\bfseries]{\#},
  emph={SYSTEM,PROMPT,USER,TEMPLATE,FOR,COMPRESSED,MEMORY,GOLD,TEXT,
        STANDARD,ANSWER,CHECKING,TEMPORAL,REASONING,VARIANT,KNOWLEDGE,
        UPDATE,PERSONALIZED,PREFERENCE,ABSTENTION,REWARD,MAPPING},
  emphstyle=\bfseries,
  frame=single,
  rulecolor=\color{black!30},
  xleftmargin=0.5em,
  xrightmargin=0.5em,
  captionpos=b
}

\newcolumntype{Y}{>{\centering\arraybackslash}X}

\title{LazyMem: Retrieve Broadly,\\
Construct Selectively for Efficient\\
Long-Term Agent Memory}

\author{
\textbf{Jing Yu\textsuperscript{1}},
\textbf{Yibo Zhao\textsuperscript{1}},
\textbf{Jiaming Zhang\textsuperscript{1}},
\textbf{Xiang Li\textsuperscript{1}\thanks{Corresponding author: \texttt{xiangli@dase.ecnu.edu.cn}}}
\\
\textsuperscript{1}School of Data Science and Engineering, East China Normal University
}

\iclrfinalcopy
\begin{document}

\maketitle
\pagestyle{fancy}
\fancyhf{}
\lhead{Preprint. Under review.}
\fancyfoot[C]{\thepage}
\renewcommand{\headrulewidth}{0.4pt}

\begin{abstract}
Long-term memory enables LLM agents to leverage past interactions, but dialogue
histories quickly exceed the context window, forcing agents to retrieve relevant
subsets at query time. Because useful evidence is sparse and scattered across
verbose conversations, retrieval faces a fundamental tension: broadening recall
improves coverage but floods downstream reasoning with noise, while compressing
memories at write time eases retrieval but irreversibly discards details that
future queries may need.
We introduce \textsc{LazyMem}, which resolves this tension by deferring all
memory construction to query time. Given a retrieved candidate pool, a
lightweight model processes it in overlapping parallel windows, selectively
retaining and compressing only query-relevant content. The model is trained with
supervised fine-tuning followed by reinforcement learning, where the reward
jointly encourages identifying relevant messages and producing compressions that
are faithful to the source and useful for answering the query.
On LongMemEval, \textsc{LazyMem}-4B achieves an LLM-judge accuracy of 0.85, outperforming the strongest non-oracle baseline while using only 213 answer-context memory tokens—21.0$\times$ fewer than the baseline.
It further generalizes to LoCoMo without target-domain training and reduces mean latency relative to the prior query-time baseline. Code is available at
\url{https://github.com/allacnobug/LazyMem}.

\end{abstract}

\section{Introduction}
\label{sec:intro}
By maintaining long-term conversational memory, LLM agents can preserve past exchanges with users and leverage them in subsequent conversations~\citep{10.1145/3586183.3606763, packer2024memgptllmsoperatingsystems, zhong2024memorybank}.
However, the accumulated dialogue history can far exceed the context window, and including it indiscriminately buries relevant evidence in noise.
The agent must therefore retrieve a query-relevant subset of its history and incorporate it into the current context. 
This retrieval step mirrors RAG in purpose but differs in the quality and structure of the underlying corpus.
Conventional RAG assumes curated, information-dense passages~\citep{NEURIPS2020_6b493230, karpukhin-etal-2020-dense}, whereas raw agent memories are loosely structured and verbose, with salient facts scattered across interaction turns~\citep{maharana2024evaluating, wu2025longmemeval}. 
This mismatch creates a dilemma. 
Noise and fragmentation make query-relevant evidence difficult to retrieve, yet expanding the recalled pool to improve evidence coverage increases context burden and further dilutes salient information~\citep{shi2023large, liu2024lost}.

Most existing methods seek to mitigate this dilemma through a \emph{construct-then-retrieve} paradigm. 
They summarize~\citep{zhong2024memorybank, fang2026lightmem}, rewrite~\citep{chhikara2025mem0, zhang2026memskill}, aggregate~\citep{sun2025hierarchical, li-etal-2026-timem}, or link raw interactions into structured memory units~\citep{xu2026structmem, xuamem} before storage, and retrieve relevant units at query time. 
By compressing and organizing raw interactions in advance, this paradigm facilitates query matching and reduces the burden on downstream reasoning.
However, because this memory construction process is necessarily query-agnostic, it may discard details that later prove relevant.
An ablation study reported in \textsc{AMA-Bench}~\citep{zhao2026ama} shows that, even when retrieval errors are excluded, models given constructed memories perform substantially worse than those given the corresponding gold interaction turns, exposing the information loss introduced during memory construction (see Appendix~\ref{app:ama_diagnostic} for details).

A natural alternative is to preserve raw interactions and defer construction to query time. 
With the query available, the system can determine which facts to retain and how aggressively to compress them, avoiding the permanent information loss inherent in query-agnostic construction.
\citet{wu2026back} provides a starting point with Query-Driven Pruning: raw conversations are retrieved, and a separate language model extracts query-relevant content through prompting before passing the resulting evidence to the answer model. 
This design demonstrates the feasibility of a \emph{retrieve-then-construct} paradigm, where raw interactions are preserved, and evidence is constructed conditioned on the query.
However, this approach has two limitations. 
First, although extraction is decoupled from answer generation, the trade-off between retrieval coverage and context burden remains. 
The extraction model still processes the entire recalled pool at once, leaving it vulnerable to context rot as salient evidence is diluted by lengthy, noisy context.
Second, prompt-based extraction introduces an additional trade-off among extraction quality, cost, and latency. Large models can achieve reasonable extraction quality without task-specific training but are costly and slow, whereas lightweight models are cheaper and faster but perform poorly without dedicated training.
These findings motivate the central question of this work: how can agent memory systems balance information fidelity against context burden under practical cost and latency constraints?

In this paper,
we introduce \textsc{LazyMem}, a \emph{retrieve-then-construct} agent memory system designed to address these trade-offs.
Its name reflects two senses of laziness: it performs no lossy construction at write time, deferring memory construction until a query arrives, and it delegates construction to a lightweight memory-processing model, freeing the main reasoning model from processing raw memory.
At query time, \textsc{LazyMem} first retrieves a large pool of raw observations to ensure adequate evidence coverage; the memory-processing model then constructs compact, query-conditioned evidence for downstream reasoning.
To control context burden, it processes the recalled pool in overlapping windows, keeping each input short enough to mitigate context rot while preserving evidence across window boundaries.
To achieve high construction quality despite its small size, we train it with reinforcement signals that jointly optimize fidelity to the original interactions and downstream reasoning utility.
The windows are processed in parallel to bound latency as the recalled pool grows, while the model's small size keeps inference cost low.

Our main contributions are summarized as follows.
\begingroup
\setlength{\emergencystretch}{2em}
\begin{itemize}

    \item We propose \textsc{LazyMem}, a retrieve-then-construct agent memory system that preserves raw interactions, retrieves broadly at query time, and constructs compact evidence through a lightweight model operating over overlapping parallel windows, avoiding irreversible write-time loss while bounding context length and latency.

\item We develop a 4B memory-processing model trained with SFT and format-gated reinforcement learning, using a joint reward that promotes selection accuracy, faithfulness, and query utility. Ablations show that SFT improves format compliance, while RL further enhances memory-construction quality, together outperforming prompting alone.

\item Across two established benchmarks, \textsc{LazyMem}-4B achieves an LLM-judge accuracy of 0.85 on LongMemEval using only 213 answer-context memory tokens, 68.7$\times$ fewer than retrieval-only and 21.0$\times$ fewer than the strongest non-oracle baseline, and reaches 0.68 on LoCoMo without target-domain training, while reducing mean end-to-end latency relative to the prior query-time method.

\end{itemize}
\endgroup

\section{Related Work}
Early efforts such as MemoryBank~\citep{zhong2024memorybank} and
MemGPT~\citep{packer2024memgptllmsoperatingsystems} equip LLM agents with persistent memory beyond finite context windows. 
Despite their different designs, these systems share a general pipeline that stores past interactions, retrieves information relevant to the current query, and incorporates the retrieved content into the model
context.
From the perspective of when past interactions are transformed into task-usable memory, we broadly organize existing approaches into two paradigms:
\emph{construct-then-retrieve} and \emph{retrieve-then-construct}.

\emph{Construct-then-retrieve} methods transform interactions into compact or structured memory representations before future queries are known. 
One line manages persistent memory through explicit operations, from prompt-based execution of a predefined set of ADD/UPDATE/DELETE operations~\citep{chhikara2025mem0} to operation policies learned via reinforcement learning~\citep{yan2026memory,yue2026memt} and reusable,
evolvable memory skills~\citep{zhang2026memskill}.
Another organizes interaction history hierarchically, consolidating concrete interactions into temporal, topical, and relational structures at varying granularities and abstraction levels~\citep{fang2026lightmem,li-etal-2026-timem,xu2026structmem}. 
On the retrieval side, mechanisms are tailored to these representations, including semantic
links among atomic notes~\citep{xuamem}, entity--relation graphs~\citep{chhikara2025mem0}, hierarchical top-down search~\citep{sun2025hierarchical}, and searched modular memory architectures~\citep{zhang2026memevolve}. 
Despite their effectiveness, their write-time construction is necessarily query-agnostic, creating a potential mismatch with unforeseen future information needs.

\begingroup
\setlength{\emergencystretch}{1.5em}
\emph{Retrieve-then-construct} methods instead retain fine-grained records close to the original interactions and defer modeling to the construction
stage, forming task-specific context by filtering, compressing, and reorganizing recalled content once the query is known. 
Although related ideas
such as reranking~\citep{yu2024rankrag}, context
compression~\citep{xu2024recomp}, and query-focused
summarization~\citep{edge2024local} are well studied in conventional RAG, this
paradigm remains underexplored in agent memory.
NanoMemory~\citep{wu2026back} preserves raw conversations and combines turn
isolation retrieval with query-driven pruning to distill recalled sessions
into compact, information-dense context. 
However, because it processes the entire recalled pool at once, context rot
remains a challenge during memory construction. Moreover, its prompt-based
memory-processing model faces a trade-off between construction quality and inference
efficiency.

Our \textsc{LazyMem} builds on this paradigm by formulating query-conditioned memory construction as a learned policy implemented by a lightweight model. It
processes large candidate pools through overlapping windows in parallel, limiting the context handled in each construction step while controlling
query-time latency.
\endgroup

\section{Method}
\label{sec:method}

\subsection{The LazyMem Pipeline}
\label{sec:pipeline}

\begin{figure}[t]
\centering
\includegraphics[width=1\textwidth]{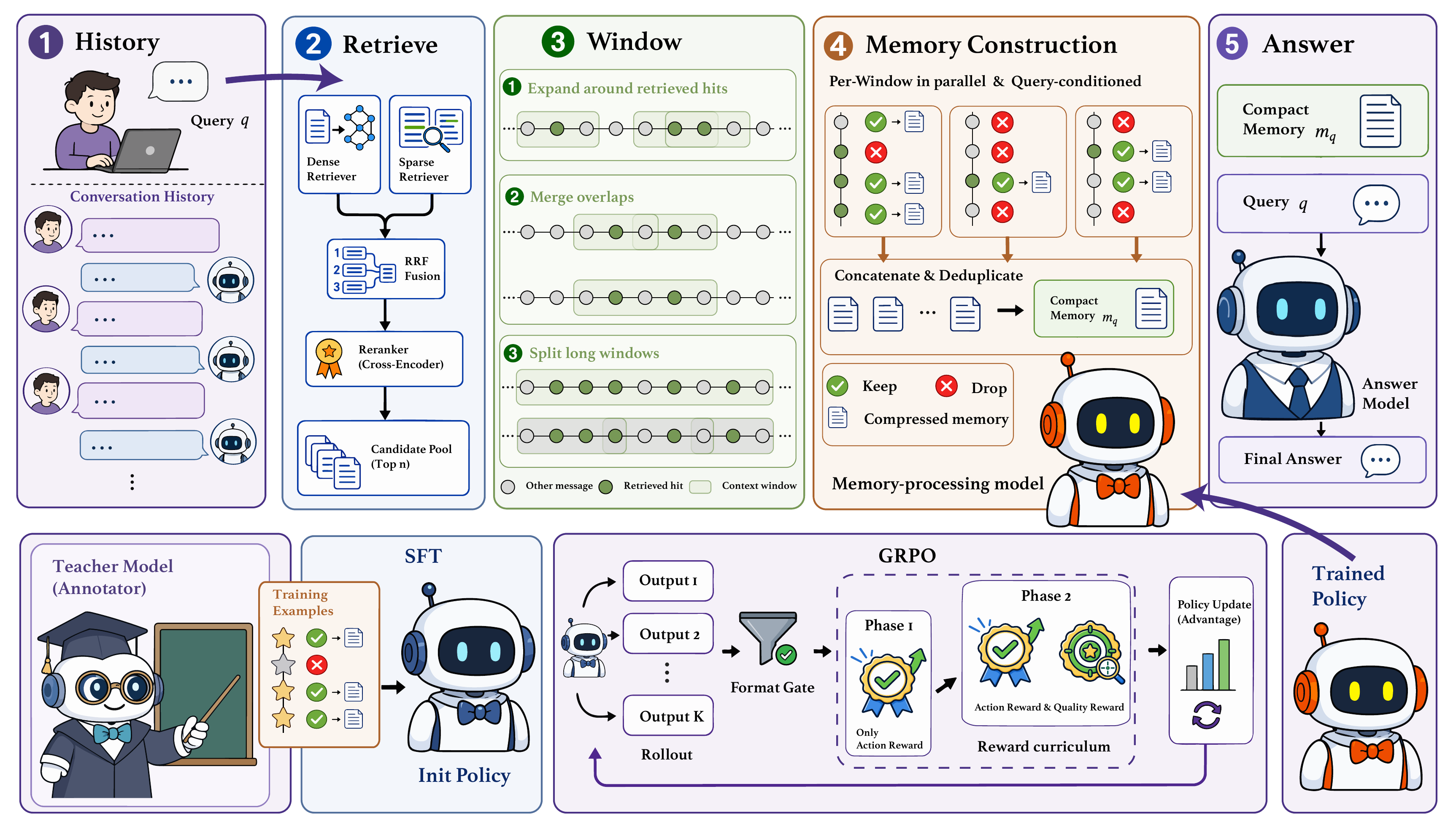}
\caption{
Overview of \textsc{LazyMem}.
At query time (top), hybrid retrieval, context windowing, and selective
construction turn a long conversation history into compact memory for the
answer model. The memory-processing model is initialized with teacher-supervised
fine-tuning and refined through reinforcement learning with action and quality
rewards (bottom).
}
\label{fig:overview}
\Description{The figure has two rows. The top row is a left-to-right inference
pipeline with four stages. Retrieve: a query and the conversation history feed a
dense retriever and a sparse retriever, whose ranked lists are combined by
Reciprocal Rank Fusion and scored by a cross-encoder reranker to form a top-n
candidate pool. Window: each retrieved hit is expanded into a surrounding context
window, overlapping windows are merged, and over-long windows are split. Memory
construction: the memory-processing model processes each window in parallel and
query-conditioned, labeling every message keep or drop, emitting compressed text
for kept messages, and then concatenating and deduplicating them into a compact
memory. Answer: the compact memory and the query are given to the answer model,
which produces the final answer. A legend distinguishes other messages, retrieved
hits, context windows, and the memory-processing model. The bottom row shows
training of the memory-processing model. A teacher model acts as annotator to
produce training examples used for supervised fine-tuning, yielding an initial
policy. This policy is then optimized with GRPO: it generates a group of rollout
outputs that pass through a format gate, are scored by a reward curriculum with
Phase 1 using only the action reward and Phase 2 adding both the action reward
and the quality reward, and the resulting advantages drive a policy update that
produces the trained policy.}
\end{figure}

We consider the problem of answering a query $q$ over a long interaction history,
where the agent must draw on relevant information scattered across past user and
assistant messages.
LazyMem addresses this as a \emph{lazy}, \emph{retrieve-then-construct} memory
system (see Figure~\ref{fig:overview}): messages are stored verbatim with no
processing at write time, and all work is deferred to query time, where the
system operates in three stages.

\paragraph{Retrieve broadly.}
LazyMem first builds a high-recall pool of at most $n$ candidate messages,
$\mathcal{R}_q$, using standard hybrid retrieval.
Dense (bi-encoder) and sparse (BM25) retrievers run in parallel on $q$, and their
ranked lists are merged via Reciprocal Rank
Fusion~\citep{10.1145/1571941.1572114}; a cross-encoder reranker then scores the
fused candidates and retains the top $n$.
All retrieval components are off-the-shelf. The contribution of LazyMem lies in
how the resulting pool is subsequently processed.

\paragraph{Construct selectively.}
A construction module then compresses $\mathcal{R}_q$ into a compact,
query-conditioned memory context $m_q$.
It partitions the pool into overlapping evidence windows and applies a
lightweight, RL-trained memory-processing model $\pi_\theta$ to all windows in
parallel (see Section~\ref{sec:construction}).
Windowing caps the number of messages processed per model call while preserving
cross-boundary evidence, and parallelism bounds latency as the pool grows.
Section~\ref{sec:training} describes how $\pi_\theta$ attains high construction
quality despite its small size.

\paragraph{Answer.}
Finally, an answer model takes the query $q$ together with the compact context
$m_q$ to produce the prediction.

\subsection{Query-Conditioned Memory Construction}
\label{sec:construction}
Construction proceeds in two steps: a \emph{history windowing} step that
restores local conversational context around each retrieved message, and a
\emph{memory-processing model} that compresses each window into
query-relevant content.

\paragraph{History windowing.}
Messages in $\mathcal{R}_q$ arrive stripped of their surrounding conversation,
so processing them in isolation can break dialogue coherence. 
Resolving a
pronoun in a retrieved message, for instance, may require the preceding turns.
LazyMem therefore reconstructs local context directly from the interaction
history.
Let $d_1, \ldots, d_{|\mathcal{R}_q|}$ denote the retrieved messages in
chronological order, and let $\mathrm{idx}(\cdot)$ return a message's position
in the history.
Each $d_i$ is expanded into a span containing itself and $w$ messages on
either side. These neighbors supply interpretive context and often recover
relevant information that retrieval missed.
Two adjacent retrieved messages are merged into a single contiguous window whenever their spans overlap:
\begin{equation}
\mathrm{idx}(d_{i+1}) - \mathrm{idx}(d_i) \leq 2w .
\label{eq:merge}
\end{equation}
Applying this rule transitively yields $M$ maximal windows with no duplicated
context.
Here, window length is measured by the number of messages.
Windows exceeding a maximum length $L$ are split into overlapping sub-windows,
bounding the number of messages processed in every model call.
We denote the final set of windows by $\{W_1, \ldots, W_P\}$ with $P \geq M$,
and evaluate the effect of $w$ in the ablation study
(Section~\ref{sec:ablation}).

\paragraph{Memory-processing model.}
The model $\pi_\theta$ processes each pair $(q, W_j)$ independently at the
granularity of individual messages.
For every message $x \in W_j$, it predicts an action
$\alpha_x \in \{\textsc{Keep}, \textsc{Drop}\}$: a kept message is rewritten
into a compressed form $\tilde{x}$ that retains only query-relevant content,
while a dropped message produces no output.
When overlapping sub-windows predict the same message, we deduplicate by $\mathrm{idx}(\cdot)$, preferring \textsc{Keep} and breaking ties by window order.
The surviving compressions are then concatenated chronologically to form
$m_q$, which is passed to the answer model together with $q$.

\subsection{Training the Memory-Processing Model}
\label{sec:training}
We train $\pi_\theta$ with supervised fine-tuning (SFT) followed by
group-based reinforcement learning (RL).

\paragraph{Supervised fine-tuning.}
We build SFT examples from source interactions whose messages carry \emph{gold}
and \emph{non-gold} labels. Gold messages are annotated by the source benchmark
as required supporting evidence; non-gold messages are unannotated, not
necessarily irrelevant, and may still provide auxiliary context.
Using the candidate retrieval and history windowing described above, we present the query and each evidence window to an instruction-following teacher model, which performs the memory-processing task: it labels every message \textsc{Keep} or \textsc{Drop} and generates query-relevant compressed text for each \textsc{Keep} message. We discard an instance if the teacher drops a gold message, produces unparsable output, violates the schema, or returns a number of decisions that does not match the window size. Each retained example maps a query and one evidence window to the per-message actions and the compressions of all \textsc{Keep} messages.
Supervised fine-tuning on these examples teaches the model the output format and basic \textsc{Keep}/\textsc{Drop} behavior, and the resulting checkpoint serves as the reference policy $\pi_{\mathrm{ref}}$. Full implementation details are reported in Section~\ref{sec:experimental_setup}.

\paragraph{Group-based policy optimization.}
Starting from $\pi_{\mathrm{ref}}$, we refine $\pi_\theta$ with Group Relative
Policy Optimization
(GRPO;~\citep{shao2024deepseekmathpushinglimitsmathematical}), modified as
below.
Each prompt $p$ pairs the query $q$ with one evidence window.
The sampling policy $\pi_{\theta_{\mathrm{old}}}$ generates $G$ candidate
outputs $\{o^k\}_{k=1}^{G}$, and each receives a scalar reward
$r^k = R(o^k;\,p)$ from the composite reward of Eq.~\ref{eq:reward}.
Each output's advantage is computed relative to its group:
\begin{equation}
\hat{A}^k
= \frac{r^k - \mathrm{mean}\{r^l\}_{l=1}^{G}}
     {\mathrm{std}\{r^l\}_{l=1}^{G} + \delta},
\label{eq:advantage}
\end{equation}
where $\delta$ is a small constant for numerical stability.
For each generated token $o_t^k$, we compute the importance-sampling ratio
between the current and the sampling policy:
\begin{equation}
\rho_t^k
= \frac{\pi_\theta(o_t^k \mid p,\; o_{<t}^k)}
     {\pi_{\theta_{\mathrm{old}}}(o_t^k \mid p,\; o_{<t}^k)} .
\label{eq:ratio}
\end{equation}
We use standard symmetric PPO clipping with clip ratio $\epsilon$, yielding
the per-token surrogate:
\begin{equation}
l_t^k = \min\!\bigl(
\rho_t^k \, \hat{A}^k,\;
\mathrm{clip}(\rho_t^k,\, 1{-}\epsilon,\, 1{+}\epsilon)\, \hat{A}^k
\bigr).
\label{eq:clip}
\end{equation}
We retain a small reference-policy KL penalty from GRPO to keep
the policy close to $\pi_{\mathrm{ref}}$, which helps preserve the output
format learned during SFT. The resulting objective is:
\begin{equation}
\mathcal{J}(\theta)
= \mathbb{E}_{\,p,\;\{o^k\}_{k=1}^{G} \sim \pi_{\theta_{\mathrm{old}}}}
\!\left[
\frac{\sum_{k=1}^{G} \sum_{t=1}^{|o^k|} l_t^k}
   {\sum_{k=1}^{G} |o^k|}
- \beta \, D_{\mathrm{KL}}\bigl(\pi_\theta \,\|\, \pi_{\mathrm{ref}}\bigr)
\right],
\label{eq:loss}
\end{equation}
which we maximize with respect to $\theta$.
Following DAPO~\citep{dapo}, we normalize the
surrogate over all tokens in the group by $\sum_k |o^k|$, rather than averaging
per rollout. This prevents longer rollouts from being down-weighted. Clipping
remains symmetric, as specified in Eq.~\ref{eq:clip}.
The KL term $D_{\mathrm{KL}}$ uses the low-variance $k_3$
estimator~\citep{shao2024deepseekmathpushinglimitsmathematical}, with $\beta$
controlling its strength.

\paragraph{Reward design.}
The composite reward combines an \emph{action reward} $R_{\mathrm{act}}$ and a
\emph{quality reward} $R_{\mathrm{qual}}$, both normalized to $[0,1]$, gated
on format validity:
\begin{equation}
R(o;\,p)
= \mathbb{I}[o \in \mathcal{V}]
\left(
(1-\lambda_{\mathrm{qual}}) R_{\mathrm{act}}
+ \lambda_{\mathrm{qual}} R_{\mathrm{qual}}
\right).
\label{eq:reward}
\end{equation}
An output is valid ($o \in \mathcal{V}$) if it is a single JSON array
matching the prescribed schema, contains exactly one decision per message in
the window, and supplies the required compressed content for each
\textsc{Keep} message; any violation yields zero reward.
The mixing weight $\lambda_{\mathrm{qual}} \in [0,1]$ and its schedule are
specified in the reward curriculum below.

\paragraph{Action reward.}
We reuse the source dataset's own annotations: messages marked as supporting
evidence for the query are \emph{gold}, and all unmarked messages are
\emph{non-gold} under the convention above.
Let $A_{\mathrm{g}}$ be the number of gold messages assigned
\textsc{Keep} divided by the total number of gold messages, and let
$A_{\mathrm{n}}$ be the number of non-gold messages assigned
\textsc{Drop} divided by the total number of non-gold messages. For a
window containing both categories, the action reward is:
\begin{equation}
R_{\mathrm{act}}
= \eta A_{\mathrm{g}} + (1-\eta) A_{\mathrm{n}},
\label{eq:action_reward}
\end{equation}
where $\eta > 0.5$ places greater weight on gold retention.
Computing accuracy within each category separately, rather than over all
messages, prevents a large majority of easy \textsc{Drop}
decisions from overwhelming a small number of critical \textsc{Keep} decisions.
The asymmetric weight reflects that dropping required evidence is irreversible
for the downstream answer model, whereas retaining additional context preserves
the evidence but introduces a context burden. If a window contains only one
category, the corresponding category accuracy is used directly. The training
value of $\eta$ and supporting diagnostics are reported in
Appendix~\ref{app:action_reward}.

\paragraph{Quality reward.}
The quality reward scores the kept set $\mathcal{K} = \{x : \alpha_x =
\textsc{Keep}\}$. It judges only the quality of what is kept, while whether
keeping was the correct action is left to $R_{\mathrm{act}}$.
An LLM judge rates each $x \in \mathcal{K}$ on two dimensions, both in
$\{0,1,2\}$ for \emph{no}, \emph{partial}, and \emph{full}: a faithfulness score
$f_x$, measuring whether the compressed content is supported by the source
message, and a utility score $u_x$, measuring its usefulness for answering the
query.
We further validate the reliability of the reward judge through a blind expert
annotation study, with the protocol and results reported in
Appendix~\ref{app:reward_judge_validation}.
The per-message quality score is:
\begin{equation}
Q(x)
= \mathbb{I}[f_x > 0]\, \frac{f_x + u_x}{4} \;\in\; [0,1] .
\label{eq:quality_score}
\end{equation}
The indicator gates on faithfulness: fully unfaithful content ($f_x = 0$)
is unsupported by or conflicts with the source message and receives zero
reward regardless of apparent utility. Partially faithful content ($f_x = 1$)
remains grounded in the source despite imprecision and thus passes the gate,
though it earns less than fully faithful content ($f_x = 2$). Beyond the gate,
faithfulness and utility contribute additively, so a shortfall on either
dimension lowers the score gradually rather than triggering a hard pass/fail
boundary. This graded signal provides GRPO's advantage normalization
(Eq.~\ref{eq:advantage}) with richer differentiation among rollouts than a
binary gate would.
Averaging $Q(x)$ over the kept messages gives the window-level reward:
\begin{equation}
R_{\mathrm{qual}}
=
\begin{cases}
0, & \mathcal{K} = \emptyset, \\[4pt]
\displaystyle
\frac{1}{|\mathcal{K}|} \sum_{x \in \mathcal{K}} Q(x),
& \text{otherwise}.
\end{cases}
\label{eq:quality_reward}
\end{equation}

\paragraph{Reward curriculum.}
The mixing weight $\lambda_{\mathrm{qual}} \in [0,1]$ keeps the composite
reward in $[0,1]$. We schedule it in two phases. In Phase~1,
$\lambda_{\mathrm{qual}}=0$, so training relies solely on the format gate and
the inexpensive rule-based action reward. This first teaches the policy correct
\textsc{Keep}/\textsc{Drop} decisions without incurring LLM-judge costs.
Once the format-validity rate on the validation set exceeds a threshold
$\tau$, training transitions to Phase~2, where a fixed
$0 < \lambda_{\mathrm{qual}} < 1$ activates the quality reward while retaining
the action reward for the remainder of training.
Deferring the quality reward until the policy already produces well-formed
outputs avoids wasting judge calls on unparsable rollouts during early
training.
The scoring rubric and judge configuration are given in
Appendix~\ref{app:quality_rubric}, the reward-mixture and curriculum settings in
Appendix~\ref{app:reward_hyper}, and the full judge prompt in
Appendix~\ref{app:reward-judge-prompt}
(Listing~\ref{lst:reward-judge-prompt}).

\section{Experiments}
\label{sec:exp}

Our experiments center on the trade-off studied in this paper: preserving
useful memory information while controlling context overhead under practical
cost and latency constraints. We organize the evaluation around four research
questions.

\textbf{RQ1 (Effectiveness):} does \textsc{LazyMem} outperform competitive memory
baselines, both in-domain and out-of-domain
(Section~\ref{sec:main-results})?

\textbf{RQ2 (Efficiency):} are these gains achieved with lower context
overhead, as measured by inference latency and answer-context memory tokens
(Section~\ref{sec:efficiency})?

\textbf{RQ3 (Ablations):} how do the history-window radius and the cumulative
SFT and RL stages affect performance
(Section~\ref{sec:ablation})?

\textbf{RQ4 (Further Analysis):} where does \textsc{LazyMem} fail, and how do
errors distribute across pipeline stages
(Section~\ref{sec:analysis})?


Complete efficiency measurements, implementation details, error-audit results,
and ethical considerations are provided in
Appendices~\ref{app:efficiency_details}, \ref{app:setup},
\ref{app:case-study}, and~\ref{app:ethics}, respectively.

\subsection{Experimental Setup}
\label{sec:experimental_setup}

\paragraph{Datasets.}
We evaluate on two long-context memory benchmarks.
\textbf{LongMemEval}~\citep{wu2025longmemeval} contains 500 questions, each
grounded in a long multi-session user--assistant history.
Because our method requires training data, we split the dataset into
360 training, 40 validation, and 100 test examples via stratified sampling over
question types.
All baselines are re-evaluated on our test split with the same answer model,
so reported numbers are directly comparable to one another, though not to
previously published full-set results.
\textbf{LoCoMo}~\citep{maharana2024evaluating} is used strictly for testing
and is excluded from both SFT and RL\@.
Following~\citet{zhang2026memskill}, we evaluate category 1--4 questions from
the last two conversations (314 questions).
LongMemEval thus measures in-domain performance and LoCoMo measures
out-of-domain generalization.

\paragraph{Training.}
We use Qwen3-4B~\citep{yang2025qwen3technicalreport} as the base
memory-processing model, and first apply supervised fine-tuning to teach it the
memory-selection and compression format.
The SFT data comes from 100 questions sampled from the LongMemEval training
split (stratified by question type) and annotated by
DeepSeek-V4-Flash~\citep{deepseekai2026deepseekv4highlyefficientmilliontoken}.
Because most messages in a window lack gold-evidence annotations, the teacher
outputs are dominated by \textsc{Drop}. We therefore balance at the window
level to reduce this action imbalance, yielding 776 SFT instances.

We then train with
GRPO~\citep{shao2024deepseekmathpushinglimitsmathematical}, using symmetric PPO
clipping and DAPO-style group-level token
normalization~\citep{dapo}. We balance RL
prompts 50/50 by whether they contain gold evidence, ensuring that gold-bearing
windows are sufficiently represented.
The reward combines the rule-based action reward $R_{\mathrm{act}}$ and the
LLM-judged quality reward $R_{\mathrm{qual}}$ under the format gate of
Eq.~\ref{eq:reward}, following the two-phase curriculum of
Section~\ref{sec:training}.

\paragraph{Baselines.}
We compare \textsc{LazyMem} against methods spanning the main memory-construction paradigms.
(i)~\emph{Oracle references}: \textit{Oracle Turn} and \textit{Oracle Session}
supply the gold evidence turn or session directly to the answer model, and serve
as reference points.
(ii)~\emph{Retrieval without construction}: \textit{RAG Top20/Top50} use the same
hybrid retrieval as \textsc{LazyMem} (Section~\ref{sec:pipeline}), then concatenate
the top results in chronological order without further processing.
(iii)~\emph{Construct-then-retrieve}: systems that build or update memory before
the query arrives. This includes the training-free
\textit{LightMem}~\citep{fang2026lightmem},
\textit{StructMem}~\citep{xu2026structmem},
\textit{Mem0}~\citep{chhikara2025mem0}, and
\textit{MemoryBank}~\citep{zhong2024memorybank}, and the trained
\textit{MemSkill}~\citep{zhang2026memskill} and
\textit{MemT}~\citep{yue2026memt}. 
(iv)~\emph{Retrieve-then-construct}: \textit{NanoMemory}~\citep{wu2026back},
which stores raw messages and constructs query-relevant context at
inference time, the same paradigm as \textsc{LazyMem}.
Unless otherwise specified, baselines use
Qwen3-32B~\citep{yang2025qwen3technicalreport} with thinking mode enabled as
the memory-processing model that constructs each system's memory; for
\textsc{LazyMem}, we report both a prompt-only Qwen3-32B variant without
task-specific training and our trained \textsc{LazyMem}-4B model.

\paragraph{Retrieval, windowing, and answer model.}
For \textsc{LazyMem}, we retrieve a candidate pool of $n{=}50$ messages with the
hybrid retrieval pipeline of Section~\ref{sec:pipeline}, using
Qwen3-Embedding-8B~\citep{zhang2025qwen3embeddingadvancingtext} for dense
embeddings, Okapi BM25%
\footnote{\url{https://github.com/dorianbrown/rank_bm25}}
for sparse retrieval, and BGE-Reranker-v2-M3%
\footnote{\url{https://huggingface.co/BAAI/bge-reranker-v2-m3}}
for reranking, and expand each message with a context window of radius $w{=}2$.
Across SFT data construction, RL prompts, and inference, we cap each sub-window
at $L{=}8$ messages and use a stride of $s{=}7$, so consecutive full
sub-windows overlap by one message; the final sub-window may be shorter.
Except where noted, all methods share the same frozen answer model, Qwen3-32B
with thinking mode enabled, so that performance differences reflect memory
quality alone.

\paragraph{Metrics.}
Our primary metric is an LLM-as-a-judge (LJ) score~\citep{zheng2023judging}.
Specifically, DeepSeek-V4-Pro compares each prediction against the reference
answer and assigns a binary correctness label.
We report the average LJ score for each dataset, along with a breakdown by
question type.
For the efficiency analysis (RQ2), we measure two quantities: the online
end-to-end per-query latency and \emph{answer-context memory tokens}, defined
as the number of tokens in the method-produced memory block supplied to the
answer model, averaged over questions. The count excludes the query and the
fixed answer-generation prompt and is computed with the Qwen3-32B tokenizer.

\subsection{Main Results (RQ1)}
\label{sec:main-results}

\begin{table}[t]
\centering
\caption{
Accuracy on LongMemEval and LoCoMo. For LongMemEval, KU, MS, SSA, SSP, SSU, and TR denote knowledge updating, multi-session reasoning, single-session assistant, single-session preference, single-session user, and temporal reasoning, respectively. For LoCoMo, MH, Temp., Open, and Single denote multi-hop, temporal, open-domain, and single-hop questions, respectively. LJ denotes the overall LLM-as-a-judge score, and MPM denotes the memory-processing model. MemSkill uses a controller trained on LoCoMo together with Qwen3-32B as its MPM, while MemT uses a 4B model trained on LoCoMo as its MPM and answer-generation model. LongMemEval therefore constitutes an out-of-domain evaluation for both methods. Oracle Session is not defined for LoCoMo. Best and second-best results among non-oracle methods are boldfaced and underlined, respectively.
}
\label{tab:main_accuracy}
\small
\setlength{\tabcolsep}{4pt}
\renewcommand{\arraystretch}{1.0}

\resizebox{\textwidth}{!}{%
\begin{tabular}{@{}llccccccc@{\hspace{8pt}}ccccc@{}}
\toprule
\multirow{2}{*}{\textbf{Method}}
&
\multirow{2}{*}{\textbf{MPM}}
&
\multicolumn{7}{c@{\hspace{8pt}}}{\textbf{LongMemEval}}
&
\multicolumn{5}{c}{\textbf{LoCoMo}}
\\
\cmidrule(lr){3-9}
\cmidrule(lr){10-14}
&
&
\textbf{KU}
&
\textbf{MS}
&
\textbf{SSA}
&
\textbf{SSP}
&
\textbf{SSU}
&
\textbf{TR}
&
\textcolor{blue}{\textbf{LJ}}
&
\textbf{MH}
&
\textbf{Temp.}
&
\textbf{Open}
&
\textbf{Single}
&
\textcolor{blue}{\textbf{LJ}}
\\
\midrule

\multicolumn{14}{c}{\textit{Gold Context}} \\

Oracle Turn
& None
& 0.88
& 0.85
& 0.91
& 0.67
& 0.93
& 0.69
& \textcolor{blue}{0.82}
& 0.67
& 0.60
& 0.60
& 0.86
& \textcolor{blue}{0.75}
\\

Oracle Session
& None
& 1.00
& 0.81
& 1.00
& 0.67
& 1.00
& 0.65
& \textcolor{blue}{0.84}
& --
& --
& --
& --
& \textcolor{blue}{--}
\\

\midrule
\multicolumn{14}{c}{\textit{Basic Retrieval}} \\

RAG Top20
& None
& 0.81
& 0.63
& \textbf{1.00}
& 0.33
& \textbf{1.00}
& 0.54
& \textcolor{blue}{0.71}
& 0.49
& 0.51
& 0.75
& \textbf{0.81}
& \textcolor{blue}{0.67}
\\

RAG Top50
& None
& 0.88
& 0.52
& \textbf{1.00}
& 0.50
& \textbf{1.00}
& 0.73
& \textcolor{blue}{0.75}
& 0.48
& 0.46
& 0.65
& \textbf{0.81}
& \textcolor{blue}{0.66}
\\

\midrule
\multicolumn{14}{c}{
\textit{Write-Time Memory Construction (Training-Free)}
} \\

LightMem
& Qwen3-32B
& 0.81
& \underline{0.81}
& 0.27
& \underline{0.67}
& \textbf{1.00}
& \underline{0.88}
& \textcolor{blue}{0.79}
& 0.38
& 0.48
& 0.40
& 0.52
& \textcolor{blue}{0.47}
\\

StructMem
& Qwen3-32B
& 0.81
& \underline{0.81}
& 0.36
& \textbf{0.83}
& \textbf{1.00}
& \textbf{0.92}
& \textcolor{blue}{0.82}
& 0.38
& 0.55
& 0.50
& 0.57
& \textcolor{blue}{0.52}
\\

Mem0
& Qwen3-32B
& 0.63
& 0.67
& 0.64
& \textbf{0.83}
& \underline{0.93}
& 0.42
& \textcolor{blue}{0.64}
& 0.42
& 0.15
& 0.60
& 0.57
& \textcolor{blue}{0.45}
\\

MemoryBank
& Qwen3-32B
& 0.88
& 0.56
& \textbf{1.00}
& 0.33
& 0.79
& 0.58
& \textcolor{blue}{0.68}
& 0.16
& 0.42
& 0.40
& 0.51
& \textcolor{blue}{0.41}
\\

\midrule
\multicolumn{14}{c}{
\textit{Write-Time Memory Construction (Trained)}
} \\

MemSkill
& \makecell[l]{Controller\\+ Qwen3-32B}
& 0.81
& 0.41
& 0.55
& 0.50
& 0.64
& 0.58
& \textcolor{blue}{0.57}
& 0.36
& 0.43
& 0.50
& 0.44
& \textcolor{blue}{0.43}
\\

MemT
& Trained 4B
& 0.63
& 0.52
& \underline{0.91}
& 0.17
& \underline{0.93}
& 0.50
& \textcolor{blue}{0.61}
& 0.36
& \textbf{0.58}
& 0.40
& 0.68
& \textcolor{blue}{0.57}
\\

\midrule
\multicolumn{14}{c}{
\textit{Query-Time Memory Construction}
} \\

NanoMemory
& Qwen3-32B
& 0.69
& 0.78
& \textbf{1.00}
& 0.33
& \textbf{1.00}
& 0.81
& \textcolor{blue}{0.80}
& 0.46
& \underline{0.57}
& \underline{0.80}
& \textbf{0.81}
& \textcolor{blue}{\underline{0.68}}
\\

\midrule
\multicolumn{14}{c}{
\textit{Query-Time Memory Construction: Our Method}
} \\

\textbf{LazyMem}
& Qwen3-32B
& \textbf{1.00}
& \textbf{0.93}
& \textbf{1.00}
& \underline{0.67}
& \textbf{1.00}
& \underline{0.88}
& \textcolor{blue}{\textbf{0.93}}
& \textbf{0.58}
& 0.55
& \textbf{0.85}
& \underline{0.78}
& \textcolor{blue}{\textbf{0.69}}
\\

\textbf{LazyMem}
& \textsc{LazyMem}-4B
& \underline{0.94}
& 0.67
& \textbf{1.00}
& \underline{0.67}
& \textbf{1.00}
& \underline{0.88}
& \textcolor{blue}{\underline{0.85}}
& \underline{0.55}
& \textbf{0.58}
& 0.75
& \underline{0.78}
& \textcolor{blue}{\underline{0.68}}
\\

\bottomrule
\end{tabular}%
}
\end{table}

Table~\ref{tab:main_accuracy} answers RQ1 affirmatively: \textsc{LazyMem}
achieves the best overall performance both in-domain and out-of-domain. On
LongMemEval, \textsc{LazyMem} with Qwen3-32B reaches an LJ of 0.93, 0.11 above
the strongest non-oracle baseline (StructMem, 0.82), and \textsc{LazyMem}-4B
still tops every baseline at 0.85. On LoCoMo, the 32B variant again leads
(0.69) and the 4B variant follows at 0.68 without any LoCoMo training data, yet
ahead of the LoCoMo-trained MemT and MemSkill. The strongest query-time methods
outperform the strongest write-time ones on both benchmarks.

\paragraph{Where the gains come from, and why they can exceed oracle context.}

The gains concentrate on aggregation- and temporal-oriented types. On KU and MS,
the 32B variant reaches 1.00 and 0.93, improving over the best baseline by 0.12
in both. On MS and TR, it even exceeds both gold-context oracles, whose own
scores stay well below saturation (Oracle Turn: 0.85/0.69; Oracle Session:
0.81/0.65). This is possible because a single oracle turn or session cannot
contain evidence scattered across multiple sessions, and providing the full
session in fact introduces noise.
Query-time construction sidesteps both problems by filtering retrieved context
conditioned on the question, removing noise while preserving cross-session
evidence for downstream reasoning. On single-fact types (KU, SSA, SSU), where
oracle context is already sufficient, both \textsc{LazyMem} and the oracles
saturate near 1.00. 
On SSP, both \textsc{LazyMem} variants match the oracle score of 0.67 but trail
the strongest write-time methods (StructMem and Mem0, both 0.83). These
questions ask for recommendations grounded in user history, a setting that may
favor persistent user profiles over on-demand evidence assembly.
We further examine how training affects this preference-sensitive construction
in Section~\ref{sec:ablation}.

The lightweight 4B variant ranks top-two on five of six LongMemEval types; the
sole exception is MS, and the overall 32B$\rightarrow$4B drop
(0.93$\rightarrow$0.85) comes almost entirely from this type
(0.93$\rightarrow$0.67). We analyze the MS gap in conjunction with the error attribution in
Section~\ref{sec:analysis}.

\subsection{Efficiency Analysis (RQ2)}
\label{sec:efficiency}

\begin{figure}[t]
\centering
\includegraphics[width=1\textwidth]{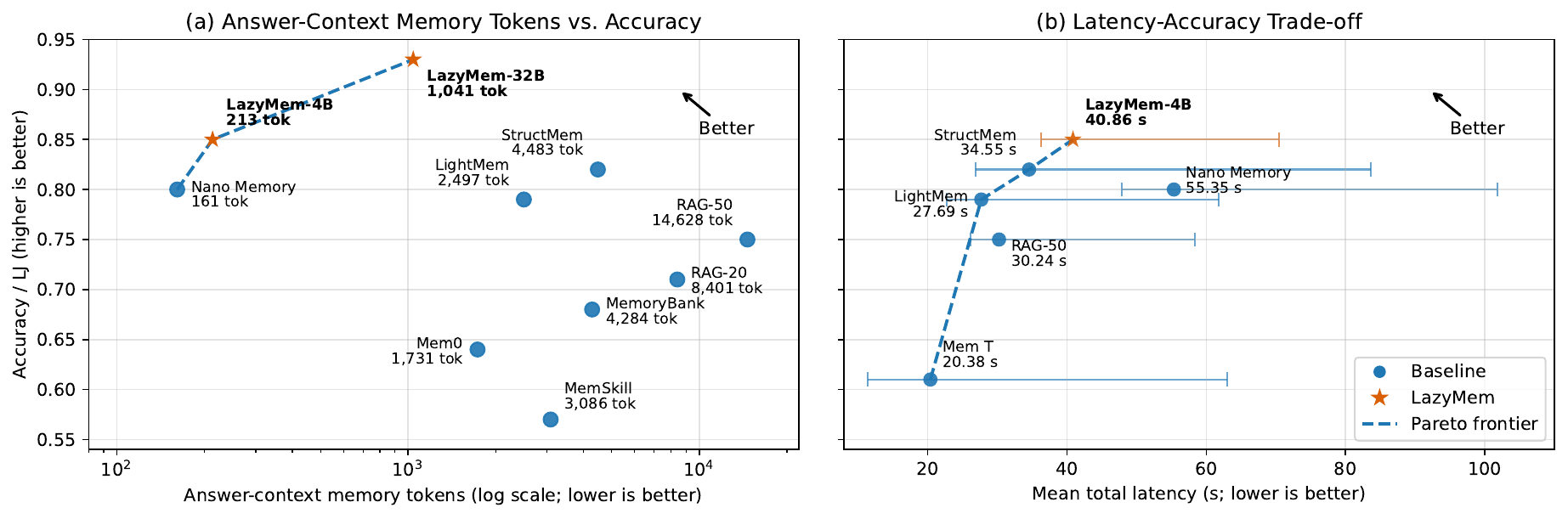}
\caption{
Accuracy--efficiency trade-offs on LongMemEval.
Left: LJ versus average answer-context memory tokens (log scale); MemT is
omitted because it lacks a comparable token count.
Right: LJ versus mean end-to-end latency under the same serving setup; bars
span P50--P95, and \textsc{LazyMem} uses the 4B memory-processing model.
Dashed lines mark Pareto frontiers (upper left is better). Labels report token
counts (left) and mean latencies (right).
}
\label{fig:efficiency_tradeoff}
\Description{Two scatter panels plot LJ accuracy on the vertical axis against an
efficiency cost on the horizontal axis, with a dashed Pareto frontier in each and
the upper-left corner being best. In the left panel the horizontal axis is the
average number of answer-context memory tokens on a log scale: both
LazyMem configurations sit on the frontier, with LazyMem-4B reaching an LJ of
about 0.85 at roughly 213 tokens and the 32B variant about 0.93 at roughly 1,041
tokens, far to the left of retrieval-only and write-time methods that use many
more tokens. In the right panel the horizontal axis is mean end-to-end latency in
seconds, with horizontal bars spanning the P50 to P95 range: LazyMem-4B lies on
the latency-accuracy frontier, achieving higher accuracy than the write-time
methods LightMem and StructMem and both higher accuracy and lower latency than
the query-time baseline NanoMemory.}
\end{figure}

Figure~\ref{fig:efficiency_tradeoff} shows that \textsc{LazyMem} offers a strong accuracy--context trade-off, with both configurations lying on the answer-context-memory-token--accuracy Pareto frontier. \textsc{LazyMem}-4B achieves an LJ of $0.85$ with an average of only 213 answer-context memory tokens, reducing this count by factors of $68.7$ relative to RAG Top50 and $21.0$ relative to StructMem, while improving LJ by $0.10$ and $0.03$, respectively. Using Qwen3-32B as the memory-processing model further increases LJ to $0.93$ with 1,041 answer-context memory tokens. These results indicate that query-conditioned construction can substantially reduce the context passed to the answer model while maintaining strong downstream accuracy.

The latency results show that \textsc{LazyMem}-4B also provides a favorable accuracy--latency trade-off. Although it has higher online latency than the write-time methods LightMem and StructMem, it achieves higher accuracy. Compared with NanoMemory, the other query-time construction method, \textsc{LazyMem}-4B improves LJ from $0.80$ to $0.85$ while reducing mean end-to-end latency from 55.35\,s to 40.86\,s, placing it on the latency--accuracy Pareto frontier. Its lightweight trained memory-processing model and parallel window processing keep the additional query-time overhead manageable.

\subsection{Ablation Studies (RQ3)}
\label{sec:ablation}

\begin{table}[H]
\centering
\small
\setlength{\tabcolsep}{5.5pt}
\renewcommand{\arraystretch}{1.08}
\caption{
Ablation of history-window radius $w$. LJ is answer accuracy, and All@50 is the
fraction of questions whose complete gold evidence is covered after augmenting
the top-50 retrieved messages with their neighboring turns. Best results in
each column are boldfaced.
}
\label{tab:window_radius}
\begin{tabular}{@{}ccccc@{}}
\toprule
\multirow{2}{*}{\textbf{Radius $w$}}
& \multicolumn{2}{c}{\textbf{LongMemEval}}
& \multicolumn{2}{c}{\textbf{LoCoMo}} \\
\cmidrule(lr){2-3}
\cmidrule(lr){4-5}
& \textbf{LJ} & \textbf{All@50}
& \textbf{LJ} & \textbf{All@50} \\
\midrule
0 & 0.77 & 0.950 & 0.63 & 0.781 \\
1 & 0.82 & 0.970 & 0.64 & 0.847 \\
2 & \textbf{0.85} & \textbf{0.980} & 0.68 & 0.889 \\
3 & \textbf{0.85} & \textbf{0.980} & \textbf{0.69} & \textbf{0.892} \\
\bottomrule
\end{tabular}
\end{table}

Table~\ref{tab:window_radius} reports the effect of history-window radius $w$.
Increasing $w$ improves evidence coverage, but LongMemEval LJ saturates at
$0.85$ for $w{\geq}2$ because All@50 is already $0.980$, leaving little room
for neighboring context to help. LoCoMo, with lower recall ($0.889$), benefits
marginally from $w{=}3$ ($0.68{\rightarrow}0.69$), though neighboring context
cannot recover evidence outside retrieved regions.

\FloatBarrier
\begin{table}[t]
\centering
\small
\setlength{\tabcolsep}{4pt}
\renewcommand{\arraystretch}{1.15}
\caption{
Cumulative training-stage ablation for the \textsc{LazyMem}-4B
memory-processing model (MPM) on LongMemEval and LoCoMo.
Qwen3-4B is the prompt-only base MPM, SFT adds supervised fine-tuning,
and RL further applies GRPO training.
Gate Fmt.\ is the fraction of outputs containing a parseable JSON
decision array with exactly one decision per input message; it measures
structural compliance rather than decision correctness.
}
\label{tab:ablation_conversational}
\resizebox{\textwidth}{!}{
\begin{tabular}{lcccccccccccccc}
\toprule
\multirow{2}{*}{\textbf{MPM}}
& \multicolumn{8}{c}{\textbf{LongMemEval}}
& \multicolumn{6}{c}{\textbf{LoCoMo}} \\
\cmidrule(lr){2-9}
\cmidrule(lr){10-15}
& \textbf{KU}
& \textbf{MS}
& \textbf{SSA}
& \textbf{SSP}
& \textbf{SSU}
& \textbf{TR}
& \textcolor{blue}{\textbf{Gate Fmt.}}
& \textcolor{blue}{\textbf{LJ}}
& \textbf{MH}
& \textbf{Temp.}
& \textbf{Open}
& \textbf{Single}
& \textcolor{blue}{\textbf{Gate Fmt.}}
& \textcolor{blue}{\textbf{LJ}} \\
\midrule

Qwen3-4B
& 0.63
& 0.19
& 0.55
& 0.17
& 0.71
& 0.35
& \textcolor{blue}{0.397}
& \textcolor{blue}{0.41}
& 0.39
& 0.45
& 0.70
& 0.66
& \textcolor{blue}{0.622}
& \textcolor{blue}{0.56} \\

Qwen3-4B (SFT)
& 0.94
& 0.56
& 1.00
& 0.17
& 1.00
& 0.65
& \textcolor{blue}{0.963}
& \textcolor{blue}{0.73}
& 0.46
& 0.52
& 0.75
& 0.74
& \textcolor{blue}{0.985}
& \textcolor{blue}{0.63} \\

Qwen3-4B (RL)
& 0.94
& 0.67
& 1.00
& 0.67
& 1.00
& 0.88
& \textcolor{blue}{0.997}
& \textcolor{blue}{0.85}
& 0.55
& 0.58
& 0.75
& 0.78
& \textcolor{blue}{0.997}
& \textcolor{blue}{0.68} \\
\bottomrule
\end{tabular}
}
\end{table}

\begin{figure}[!t]
    \centering
    \includegraphics[width=1\textwidth,trim=14.1bp 4.9bp 33.5bp 7.0bp,clip]{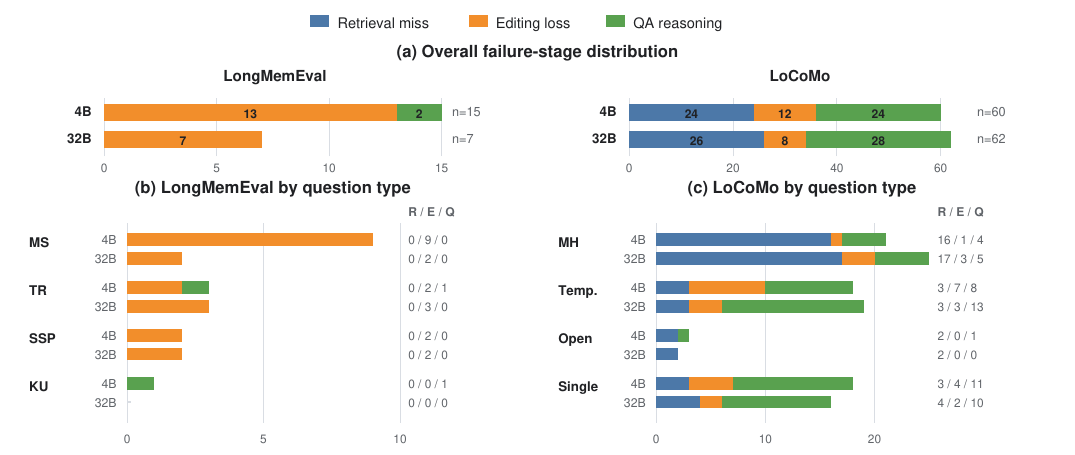}
    \caption{
    Human-attributed errors for \textsc{LazyMem}-4B and the prompt-only 32B
    variant, overall and by question type. Triplets denote retrieval miss /
    editing loss / QA reasoning (R/E/Q). SSA and SSU have zero errors.
    }
    \Description{Three stacked-bar panels compare retrieval misses, editing
    losses, and QA reasoning errors for the 4B and 32B LazyMem variants. The
    overall panel shows editing loss dominating LongMemEval and retrieval plus
    QA errors dominating LoCoMo; two detailed panels break counts down by
    question type.}
    \label{fig:error_attribution}
\end{figure}

Table~\ref{tab:ablation_conversational} shows that cumulative training consistently improves over prompting alone. The final RL model raises LJ from 0.41 to 0.85 on LongMemEval and from 0.56 to 0.68 on LoCoMo. SFT accounts for most of the initial gain (LJ +0.32 / +0.07), largely by eliminating malformed gate outputs (Gate Fmt.\ jumps from 0.397 to 0.963 and 0.622 to 0.985). RL adds a further +0.12 / +0.05 with negligible format change, indicating that its gains come from improved memory-construction decisions rather than format compliance. This validates the staged design: SFT establishes reliable gate execution, RL refines what to retain.

Complementing the cross-method comparison in
Section~\ref{sec:main-results}, SSP shows a distinct stage-wise pattern:
accuracy remains at 0.17 after SFT but rises to 0.67 after RL. Because these
questions require converting user history into recommendation-relevant
evidence, success may depend not only on retaining relevant evidence but also on how
it is compressed. The utility reward may therefore improve query-time memory construction by making the compressed preferences more useful for the
current recommendation. Given the relatively small SSP subset, this explanation
remains tentative.

\subsection{Further Analysis (RQ4)}
\label{sec:analysis}

Three expert annotators assign each confirmed error to its first failing stage
by majority vote: a \emph{retrieval miss} when
gold evidence never enters the candidate set, an \emph{editing loss} when it is
dropped or distorted during construction, or a \emph{QA reasoning error} when
the answer model fails despite sufficient memory. For this attribution analysis
only, we exclude human-confirmed judge noise and samples with erroneous gold
annotations; consequently, the attributed error totals need not equal the error
counts implied by raw LJ.

\paragraph{Overall distribution.}

On LongMemEval, All@50 is 0.98; within the retained attribution set, there are
no retrieval misses. Editing accounts for 13/15 4B errors and all seven 32B
errors; the other two 4B errors arise in QA. The total error count falls from
15 to 7, localizing
the benefit of scaling to evidence editing rather than retrieval or downstream
answering. On LoCoMo (All@50\,=\,0.89), retrieval and QA account for 24/60
errors each for 4B, and 26/62 and 28/62 for 32B. Although scaling reduces
editing losses from 12 to 8, retrieval and QA remain the dominant bottlenecks:
multi-hop questions are primarily retrieval-limited, whereas temporal and
single-hop questions more often fail during editing or QA.

\paragraph{The 4B multi-session gap.}

The 32B$\rightarrow$4B drop is concentrated on MS (0.93$\rightarrow$0.67).
With retrieval fixed, editing losses rise from 2 to 9. The failures are not
binary \textsc{Keep}/\textsc{Drop} errors but losses within compression: the 4B
model retains the correct messages yet omits details that individually appear
minor but prove necessary when the answer model must aggregate evidence across
sessions. The 32B variant, with greater generation capacity, produces more
comprehensive compressions and avoids this failure mode. 
RL likewise yields a smaller gain on MS than on TR, where
answer-critical details are more salient and the existing rewards provide
clearer guidance for selection and compression. On MS, however, a compression
can be faithful and useful at the message level yet still omit details needed
for cross-session aggregation, indicating that these signals alone do not
ensure evidence completeness.

\section{Conclusion}
This work shows that the \emph{when} of memory construction matters as much as
the \emph{how}. \textsc{LazyMem} reframes the central design question from
\emph{what to store} to \emph{what to surface}, deferring all lossy
transformation to query time so that every retention decision is conditioned
on downstream need. This turns the recall--noise trade-off from a fixed
architectural commitment into a per-query optimization.
Our experiments demonstrate that this deferral is both effective and practical.
On LongMemEval, \textsc{LazyMem}-4B achieves an LLM-judge accuracy of 0.85 with
only 213 answer-context memory tokens, outperforming RAG Top50 by 0.10 while
using 68.7$\times$ fewer tokens. It also transfers to LoCoMo without
target-domain training and provides a stronger accuracy--latency trade-off than
the prior query-time construction baseline.


\section{Ethical Considerations}
\label{app:ethics}

Our work uses only synthetic and publicly released conversational benchmarks; no
real user data is collected or stored. The memory construction model inherits
biases present in source histories and in the LLMs used for annotation and
reward judging. All human annotation was performed by compensated expert
annotators informed of the research purpose. Because raw histories may contain
sensitive information in real deployments, systems should obtain user consent,
enforce access controls and retention limits, and support deletion before
storing or processing memories.


\bibliography{custom}
\bibliographystyle{iclr2026_conference}

\clearpage
\appendix
\section{AMA-Bench Needle Diagnostic}
\label{app:ama_diagnostic}
To understand how a memory-augmented system loses information, we report the needle diagnostic introduced by \citet{zhao2026ama}. Table~\ref{tab:ama_diagnostic} reproduces the results reported in the original work; we do not rerun the corresponding experiments. The diagnostic evaluates the same needle-containing examples under three input conditions that differ only in the information exposed to the answering model: the original observations containing the needle, the memories constructed from those observations, and the memories returned by the complete retrieval pipeline. The decrease from \textit{Full Obs. w/ Needle} to \textit{Constructed Mem. w/ Needle} measures information loss introduced during memory construction, before retrieval is involved. This construction loss is substantial for all four methods, ranging from $19.6\%$ for HippoRAG2 to $41.3\%$ for MemoryBank, indicating that relevant evidence may be omitted, compressed, or altered when raw observations are converted into persistent memories.

\begin{table}[H]
\centering
\small
\setlength{\tabcolsep}{3pt}
\caption{
Needle diagnostic results on AMA-Bench, reproduced from \citet{zhao2026ama}. \textit{Full Obs. w/ Needle} provides the answering model with the original observation turns containing the target evidence. \textit{Constructed Mem. w/ Needle} instead provides the method-specific memories constructed from those observations, thereby isolating information loss during memory construction. \textit{End-to-End System} additionally applies each method's retrieval procedure, capturing further loss due to retrieval. Percentages in parentheses denote the relative decrease from the immediately preceding column.
}
\label{tab:ama_diagnostic}
\begin{tabularx}{\columnwidth}{@{}lYYY@{}}
\toprule
\textbf{Method} &
\textbf{Full Obs. w/ Needle} &
\textbf{Constructed Mem. w/ Needle} &
\textbf{End-to-End System} \\
\midrule
HippoRAG2  & 0.46 & 0.37\newline{\scriptsize ($\downarrow 19.6\%$)} & 0.21\newline{\scriptsize ($\downarrow 43.2\%$)} \\
Mem1       & 0.46 & 0.29\newline{\scriptsize ($\downarrow 37.0\%$)} & 0.20\newline{\scriptsize ($\downarrow 31.0\%$)} \\
A-Mem      & 0.46 & 0.29\newline{\scriptsize ($\downarrow 37.0\%$)} & 0.24\newline{\scriptsize ($\downarrow 17.2\%$)} \\
MemoryBank & 0.46 & 0.27\newline{\scriptsize ($\downarrow 41.3\%$)} & 0.26\newline{\scriptsize ($\downarrow 3.7\%$)} \\
\bottomrule
\end{tabularx}

\end{table}

\FloatBarrier

\section{Effect of Retrieval Depth and Model Scale in NanoMemory}
\label{app:nano_ab}

Tables~\ref{tab:nano_longmemeval_topk} and~\ref{tab:nano_locomo_topk} study the
effects of retrieval depth and construction model size in NanoMemory on
LongMemEval and LoCoMo, respectively.
The LongMemEval latency measurements follow the common protocol in
Section~\ref{app:latency}. NanoMemory processes the entire retrieved session
pool in one construction call, which cannot be split into independent parallel
calls without changing the method; it therefore uses batch size 1 and
concurrency 1.

\begin{table}[t]
\centering
\scriptsize
\setlength{\tabcolsep}{2.8pt}
\caption{
Effect of retrieval depth and memory-construction model scale in NanoMemory on
LongMemEval.
Qwen3-32B is fixed as the answer model, while \emph{Constructor}
denotes the model used for query-conditioned memory construction.
\emph{Tok.} is the average number of retrieved tokens passed to the
memory-construction model, computed with the Qwen3-32B tokenizer.
Latency reports only memory filtering and construction, excluding retrieval
and answer generation.
}
\label{tab:nano_longmemeval_topk}

\resizebox{\textwidth}{!}{%
\begin{tabular}{clrrrrrrrrrrrr}
\toprule
\multicolumn{10}{c}{\textbf{LongMemEval}}
& \multicolumn{4}{c}{\textbf{Filtering Latency (s)}} \\
\cmidrule(lr){1-10}
\cmidrule(lr){11-14}
Top-$k$ & Constructor & Tok. & KU & MS & SSA & SSP & SSU & TR & LJ
& Mean & Median & P90 & P95 \\
\midrule
\multirow{2}{*}{3}
& Qwen3-4B
& \multirow{2}{*}{7808}
& 0.63 & 0.44 & 0.82 & 0.33 & 0.71 & 0.58 & 0.58
& 17.55 & 8.61 & 28.74 & 43.06 \\
& Qwen3-32B
&
& 0.88 & 0.56 & 1.00 & 0.33 & 0.79 & 0.69 & \textbf{0.71}
& 26.64 & 21.66 & 46.08 & 59.72 \\
\midrule

\multirow{2}{*}{5}
& Qwen3-4B
& \multirow{2}{*}{12167}
& 0.75 & 0.48 & 1.00 & 0.50 & 0.86 & 0.58 & 0.66
& 16.95 & 9.81 & 26.07 & 39.56 \\
& Qwen3-32B
&
& 0.75 & 0.52 & 1.00 & 0.50 & 0.86 & 0.73 & \textbf{0.71}
& 38.14 & 24.80 & 58.60 & 67.61 \\
\midrule

\multirow{2}{*}{10}
& Qwen3-4B
& \multirow{2}{*}{23841}
& 0.69 & 0.52 & 1.00 & 0.33 & 0.86 & 0.54 & 0.64
& 19.54 & 12.50 & 31.90 & 43.63 \\
& Qwen3-32B
&
& 0.69 & 0.78 & 1.00 & 0.33 & 1.00 & 0.81 & \textbf{0.80}
& 40.74 & 32.54 & 67.71 & 87.49 \\
\bottomrule
\end{tabular}
}
\end{table}

\begin{table}[!t]
\centering
\caption{
Effect of retrieval depth and memory-construction model scale in NanoMemory on
LoCoMo. Qwen3-32B is fixed as the answer model. LoCoMo uses larger top-$k$
values because its sessions are shorter; Top-30 has a token budget comparable
to LongMemEval Top-10.
}
\label{tab:nano_locomo_topk}
\small
\setlength{\tabcolsep}{2.8pt}
\renewcommand{\arraystretch}{1.05}
\resizebox{\columnwidth}{!}{%
\begin{tabular}{clrrrrrr}
\toprule
Top-$k$ & Constructor & Tok. & Multi & Temp. & Open & Single & LJ \\
\midrule
\multirow{2}{*}{3}
& Qwen3-4B  & \multirow{2}{*}{2646}  & 0.23 & 0.20 & 0.45 & 0.64 & 0.45 \\
& Qwen3-32B &                       & 0.29 & 0.38 & 0.60 & 0.68 & \textbf{0.53} \\
\midrule
\multirow{2}{*}{5}
& Qwen3-4B  & \multirow{2}{*}{4281}  & 0.28 & 0.29 & 0.65 & 0.70 & 0.52 \\
& Qwen3-32B &                       & 0.33 & 0.51 & 0.70 & 0.73 & \textbf{0.59} \\
\midrule
\multirow{2}{*}{10}
& Qwen3-4B  & \multirow{2}{*}{8260}  & 0.39 & 0.42 & 0.70 & 0.76 & 0.60 \\
& Qwen3-32B &                       & 0.46 & 0.57 & 0.80 & 0.81 & \textbf{0.68} \\
\midrule
\multirow{2}{*}{15}
& Qwen3-4B  & \multirow{2}{*}{12084} & 0.42 & 0.37 & 0.65 & 0.78 & 0.61 \\
& Qwen3-32B &                       & 0.46 & 0.60 & 0.80 & 0.83 & \textbf{0.70} \\
\midrule
\multirow{2}{*}{20}
& Qwen3-4B  & \multirow{2}{*}{15888} & 0.42 & 0.43 & 0.50 & 0.84 & 0.64 \\
& Qwen3-32B &                       & 0.57 & 0.54 & 0.85 & 0.82 & \textbf{0.71} \\
\midrule
\multirow{2}{*}{25}
& Qwen3-4B  & \multirow{2}{*}{19579} & 0.36 & 0.35 & 0.60 & 0.78 & 0.59 \\
& Qwen3-32B &                       & 0.52 & 0.58 & 0.85 & 0.87 & \textbf{0.73} \\
\midrule
\multirow{2}{*}{30}
& Qwen3-4B  & \multirow{2}{*}{21512} & 0.35 & 0.29 & 0.60 & 0.76 & 0.56 \\
& Qwen3-32B &                       & 0.55 & 0.54 & 0.80 & 0.87 & \textbf{0.73} \\
\bottomrule
\end{tabular}
}
\end{table}

Increasing top-$k$ improves evidence coverage but also introduces more
tokens for the construction model to filter. On LongMemEval, the 32B model
improves from $0.71$ at Top-3 to $0.80$ at Top-10, whereas the 4B model
peaks at Top-5 and then declines. A similar pattern appears on LoCoMo:
the 32B model improves up to $0.73$, while the 4B model peaks at Top-20
and degrades under deeper retrieval. This suggests that the larger
construction model handles longer and noisier retrieved contexts more
effectively.

The improved filtering quality comes at a higher computational cost.
On LongMemEval, the 32B model requires $26.64$--$40.74$ seconds of
filtering latency on average, compared with $16.95$--$19.54$ seconds for
the 4B model. The performance gap is largest at Top-10, where the retrieved
context is longest. NanoMemory therefore exhibits a clear trade-off among
retrieval coverage, filtering quality, context size, and latency.

\section{Additional Efficiency Results}
\label{app:efficiency_details}

\subsection{Answer-Context Memory Tokens}
\label{app:answer_context_tokens}
Table~\ref{tab:answer_context_memory_tokens} reports the exact
answer-context memory-token counts underlying the LongMemEval results in
Figure~\ref{fig:efficiency_tradeoff}. \textsc{LazyMem}-4B supplies only 213
tokens, $68.7\times$ fewer than RAG Top50 and $21.0\times$ fewer than
StructMem, while achieving higher LJ accuracy. The prompt-only 32B variant
uses 1,041 tokens and reaches an LJ of 0.93, illustrating the trade-off
between construction capacity and context efficiency. The LoCoMo counts are
included for completeness and show the same qualitative difference between
the two \textsc{LazyMem} configurations.

\begin{table}[t]
\centering
\caption{
Answer-context memory tokens, defined as the average number of tokens in the
method-produced memory block supplied to the answer model per question. Counts
exclude the query and the fixed answer-generation prompt and are computed with
the Qwen3-32B tokenizer.
The LongMemEval column underlies the answer-context-memory-token--accuracy
results in Figure~\ref{fig:efficiency_tradeoff}; the LoCoMo column is included
for completeness. N/A indicates that the method does
not expose an answer-context memory-token count that is directly comparable
under our accounting protocol.
}
\label{tab:answer_context_memory_tokens}
\small
\setlength{\tabcolsep}{4.5pt}
\renewcommand{\arraystretch}{1.08}

\begin{tabular}{@{}llrr@{}}
\toprule
\textbf{Method}
& \textbf{Constructor}
& \multicolumn{1}{c}{\textbf{LongMemEval}}
& \multicolumn{1}{c}{\textbf{LoCoMo}} \\
\midrule

\multicolumn{4}{c}{\textit{Gold Context}} \\
\midrule
Oracle Turn
& None
& 515
& 171 \\

Oracle Session
& None
& 5,778
& -- \\

\midrule
\multicolumn{4}{c}{\textit{Basic Retrieval}} \\
\midrule
RAG Top20
& None
& 8,401
& 1,658 \\

RAG Top50
& None
& 14,628
& 3,834 \\

\midrule
\multicolumn{4}{c}{
\textit{Write-Time Construction (Training-Free)}
} \\
\midrule
LightMem
& Qwen3-32B
& 2,497
& 2,445 \\

StructMem
& Qwen3-32B
& 4,483
& 5,010 \\

Mem0
& Qwen3-32B
& 1,731
& 8,881 \\

MemoryBank
& Qwen3-32B
& 4,284
& 1,599 \\

\midrule
\multicolumn{4}{c}{
\textit{Write-Time Construction (Trained)}
} \\
\midrule
MemSkill
& \makecell[l]{Trained Controller\\+ Qwen3-32B}
& 3,086
& 2,358 \\

MemT
& Trained 4B
& N/A
& N/A \\

\midrule
\multicolumn{4}{c}{
\textit{Query-Time Construction}
} \\
\midrule

NanoMemory
& Qwen3-32B
& 161
& 133 \\

\midrule
\multicolumn{4}{c}{
\textit{Query-Time Construction: Our Method}
} \\
\midrule
\textbf{LazyMem}
& Qwen3-32B
& 1,041
& 1,945 \\

\textbf{LazyMem}
& \textsc{LazyMem}-4B
& 213
& 697 \\

\bottomrule
\end{tabular}
\end{table}

\subsection{Latency Protocol and Breakdown}
\label{app:latency}
All latency experiments run on a server with two NVIDIA A800 80GB PCIe GPUs
and AMD EPYC 7763 CPUs. We serve LLM inference with vLLM on one GPU using
bfloat16 precision and a GPU memory utilization of 0.9, and compute question
embeddings on the other. Each method uses its own configured model size.

We process questions strictly sequentially, without batching or parallelism
across questions. For the current question, \textsc{LazyMem} issues independent
history-window requests concurrently (up to 64 at a time). NanoMemory instead
processes the entire retrieved pool in one call and cannot expose comparable
intra-query parallelism without changing its pipeline. The latency results thus
include this structural advantage of \textsc{LazyMem}. We discard three warm-up
runs and report statistics over the next three runs.

Table~\ref{tab:latency_results} breaks end-to-end latency down into
retrieval, memory processing, and answer generation. Query-time methods add
a memory-processing stage. The \textsc{LazyMem} row uses the trained 4B
memory-processing model, whereas NanoMemory uses Qwen3-32B, matching the
configurations compared in the main efficiency analysis.

\begin{table}[t]
\centering
\scriptsize
\caption{
End-to-end latency breakdown on LongMemEval.
Total latency includes retrieval, method-specific memory processing, and
answer generation. Offline memory-construction costs for write-time
methods are not included. Processing denotes memory filtering or
construction performed after retrieval. Med., P90, and P95 denote the
median, 90th percentile, and 95th percentile, respectively. A dash
indicates that the corresponding stage is not applicable.
}
\label{tab:latency_results}
\setlength{\tabcolsep}{3.2pt}
\renewcommand{\arraystretch}{1.08}
\resizebox{\textwidth}{!}{
\begin{tabular}{lrrrrrrrrrrrrrrrr}
\toprule
\multirow{2}{*}{Method}
& \multicolumn{4}{c}{Total Latency (s)}
& \multicolumn{4}{c}{Retrieval Latency (s)}
& \multicolumn{4}{c}{Processing Latency (s)}
& \multicolumn{4}{c}{Answer Latency (s)} \\
\cmidrule(lr){2-5}
\cmidrule(lr){6-9}
\cmidrule(lr){10-13}
\cmidrule(lr){14-17}
& Mean & Med. & P90 & P95
& Mean & Med. & P90 & P95
& Mean & Med. & P90 & P95
& Mean & Med. & P90 & P95 \\
\midrule
RAG Top50
& 30.24 & 26.13 & 50.99 & 58.36
& 1.58 & 1.53 & 1.89 & 1.97
& -- & -- & -- & --
& 28.66 & 24.35 & 49.41 & 56.45 \\

MemT
& 20.38 & 11.35 & 44.48 & 63.03
& 14.58 & 7.53 & 31.68 & 35.54
& -- & -- & -- & --
& 5.80 & 3.46 & 9.77 & 17.37 \\

LightMem
& 27.69 & 22.73 & 49.69 & 61.76
& 0.04 & 0.02 & 0.07 & 0.08
& -- & -- & -- & --
& 27.65 & 22.67 & 49.67 & 61.73 \\

StructMem
& 34.55 & 26.87 & 69.14 & 83.58
& 0.04 & 0.03 & 0.04 & 0.05
& -- & -- & -- & --
& 34.51 & 26.84 & 69.08 & 83.55 \\

NanoMemory
& 55.35& 47.85 & 89.39 & 101.86
& 0.05 & 0.05 & 0.06 & 0.06
& 40.74 & 32.54 & 67.71 & 87.49
& 14.56 & 12.67 & 22.01 & 26.84 \\

\textsc{LazyMem}-4B
& 40.86 & 36.31 & 61.86 & 70.49
& 1.58 & 1.53 & 1.89 & 1.97
& 23.09 & 20.83 & 36.26 & 41.83
& 16.18 & 12.78 & 28.84 & 36.70 \\
\bottomrule
\end{tabular}
}
\end{table}

\section{Experimental Setup Details}
\label{app:setup}

\subsection{Dataset and Split Construction}
\label{app:split}

LongMemEval~\citep{wu2025longmemeval} contains 500 questions, each grounded in a
long, multi-session user--assistant interaction history. The dataset defines
six question types: single-session user, single-session assistant,
single-session preference, multi-session, temporal reasoning, and knowledge
update. Abstention questions are indicated by the suffix \texttt{\_abs} in
their question IDs and retain their original question types.

Because our method requires training data, we stratify the 500 questions by
type and split them into 360/40/100 training, validation, and test examples
(72\%/8\%/20\%). Within each type, examples are sorted by question ID and
shuffled with a fixed seed of 42, while the largest-remainder method is used
to preserve the original type distribution.

Table~\ref{tab:split-stats} reports the resulting per-type counts. The test
split is excluded from SFT and RL training, hyperparameter tuning, and
checkpoint selection. All model-selection decisions are based on the
validation split. The test split is used only for final evaluation.

\begin{table}[htbp]
\centering
\caption{Question-type statistics for the LongMemEval splits.}
\label{tab:split-stats}
\begin{tabular}{lrrrr}
\toprule
Question type & Total & Train & Validation & Test \\
\midrule
Single-session user       & 70  & 50 & 6  & 14 \\
Single-session assistant  & 56  & 40 & 5  & 11 \\
Single-session preference & 30  & 22 & 2  & 6  \\
Multi-session             & 133 & 96 & 10 & 27 \\
Temporal reasoning        & 133 & 96 & 11 & 26 \\
Knowledge update          & 78  & 56 & 6  & 16 \\
\midrule
Total                     & 500 & 360 & 40 & 100 \\
\bottomrule
\end{tabular}
\end{table}

\subsection{Training Data}

Here, \emph{gold} denotes benchmark-annotated required evidence; \emph{non-gold}
denotes unannotated messages, which are not necessarily irrelevant and may
still provide auxiliary context.

\paragraph{SFT data construction.}
\label{app:sft-data}
  
The 100 questions, sampled by question type from the 360-question training
split, yield 3,075 query--window pairs. We prompt
DeepSeek-V4-Flash~\citep{deepseekai2026deepseekv4highlyefficientmilliontoken} in
high-thinking mode with the prompt in
Listing~\ref{lst:window-annotation-prompt}. For each message in a window, the
teacher predicts a keep/drop decision and, when keeping, a compressed rewrite;
its reasoning and structured decisions form the SFT target.

Since gold evidence is sparse within a window, the raw annotations are 3.92\%
keep and 96.08\% drop. We discard 36 windows that
drop gold evidence, then balance at the window level by pairing the 388 windows
containing keep decisions with 388 sampled all-drop windows, giving 776
instances with average input and target lengths of 2,585 and 873 tokens.
Automated validation finds no formatting errors in the raw annotations.

\paragraph{RL data balancing.}
\label{app:rl-data}
The raw RL training pool contains 11,072 query--window prompts, of which 8.98\%
contain gold evidence and 91.02\% do not. We construct a balanced training set of
10,000 prompts by sampling 5,000 from each group: prompts containing gold
evidence with replacement, and the others without replacement.

\subsection{Training Hyperparameters}
\label{app:training}

Table~\ref{tab:hyperparams} lists the explicitly set configuration of both
training stages. SFT uses LLaMA-Factory~\citep{zheng-etal-2024-llamafactory}
0.9.3 for full-parameter fine-tuning and runs with DeepSpeed ZeRO-2. The RL
stage uses GRPO~\citep{shao2024deepseekmathpushinglimitsmathematical}
implemented in VERL~\citep{Sheng_2025} 0.7.1 with Ray and FSDP. Both stages
train Qwen3-4B~\citep{yang2025qwen3technicalreport} on one node using two
NVIDIA A800 80GB PCIe GPUs. SFT takes
approximately 0.36 hours. 
The RL dataloader drops the final incomplete batch in each epoch, so the
10,000-prompt training set, global batch size of 32, and three epochs give
$\lfloor 10{,}000/32\rfloor\times 3=936$ optimization steps. The observed
wall-clock time is approximately 7.5--8.5 days per uninterrupted RL run,
including periodic validation and checkpointing.

\begin{table}[ht]
\centering
\caption{Explicit training hyperparameters for SFT and RL stages.}
\label{tab:hyperparams}
\small
\setlength{\tabcolsep}{1.3pt}
\begin{tabular}{lcc}
\toprule
 & SFT & RL (GRPO) \\
\midrule
Base model & Qwen3-4B & SFT checkpoint \\
Framework & LLaMA-Factory 0.9.3 & VERL 0.7.1 \\
Learning rate & $5\times10^{-6}$ & $1\times10^{-6}$ \\
LR schedule & cosine & constant \\
Warmup ratio & 0.03 & 0.0 \\
Epochs & 2 & 3 \\
Global batch size & 16 & 32 \\
Dataloader \texttt{drop\_last} & --- & Yes \\
Max input length & 8192 & 8192 \\
Max generation length & --- & 3072 \\
Rollouts per prompt & --- & 8 \\
Sampling temperature (rollout) & --- & 0.9 \\
KL coefficient & --- & 0.001 \\
KL loss type & --- & \texttt{low\_var\_kl} \\
Symmetric PPO clip ratio $\epsilon$ & --- & 0.2 \\
Precision & bf16 & bf16 \\
Optimizer & AdamW & AdamW \\
Weight decay & --- & 0.01 \\
Adam $\beta$ values & --- & $(0.9, 0.999)$ \\
Gradient clipping & --- & 1.0 \\
Distributed & DeepSpeed ZeRO-2 & Ray + FSDP \\
Rollout backend & --- & vLLM 0.14.0 (TP=1) \\
Other GRPO parameters & --- & VERL 0.7.1 defaults \\
Hardware & \multicolumn{2}{c}{$2\times$ A800 80GB PCIe} \\
\bottomrule
\end{tabular}
\end{table}

For SFT, the effective global batch size is computed as per-device batch size
$1$ times gradient accumulation steps $8$ times $2$ GPUs. For RL, the batch size
denotes the number of prompts per training batch; each prompt is sampled with
8 rollouts. The RL actor uses PPO mini-batch size 16 and per-GPU micro-batch
size 4. Rollout generation uses nucleus sampling with top-$p=0.95$.

The training node exposes eight A800 GPUs, but both stages set
\texttt{CUDA\_VISIBLE\_DEVICES=0,1} and use two GPUs per node (81,920 MiB
each). The RL actor uses FSDP with bf16 parameters and no
parameter or optimizer offloading. The reference model uses FSDP parameter
offloading and disables Torch compilation. Ulysses sequence parallelism and
vLLM tensor parallelism are both set to 1. DeepSpeed and ZeRO are not used in
the RL stage.

The training script does not override a single global random seed. Under the
resolved VERL configuration, the FSDP seed and actor dataloader seed are both
42, while \texttt{data.seed} remains unset. We validate before training and
every 10 optimization steps, save a checkpoint every 30 steps, retain at most
five actor checkpoints, and save the model, optimizer, and auxiliary training
state. All remaining RL/GRPO parameters use the defaults of VERL 0.7.1.

\subsection{Implementation and Inference Details}
\label{app:impl}

The constructed memory is concatenated with the query and passed to the answer
model; the answer prompt is shown in
Listing~\ref{lst:answer-generation-prompt}.

\paragraph{Baseline implementations.}
We implement each baseline following its original or official pipeline, while
standardizing the backbone model and retrieval budget where appropriate.

\begin{itemize}

\item \textbf{Oracle Turn and Oracle Session}.
The answer model is given the gold evidence turn or the complete session
containing it, respectively.

\item \textbf{RAG Top20 and RAG Top50}.
We retrieve the top-20 or top-50 messages per query using the same hybrid
retrieval pipeline as \textsc{LazyMem} (Section~\ref{sec:pipeline}), then
concatenate them in chronological order before answer generation.

\item \textbf{LightMem}~\citep{fang2026lightmem}.
We use the original memory-entry construction and retrieval pipeline with hybrid
user--assistant messages, retrieving the top-50 memory entries.

\item \textbf{StructMem}~\citep{xu2026structmem}.
We use the original pipeline with both structured summaries and memory entries,
adopting hybrid user--assistant messages and retrieving the top-50 memory entries
and top-5 structured summaries.

\item \textbf{Mem0}~\citep{chhikara2025mem0}.
We use the open-source Mem0 pipeline with Qwen3-Embedding-8B for dense retrieval
and Qdrant as the vector store, retrieving the top-50 memories per query and
otherwise following the official Mem0 benchmark configuration.

\item \textbf{MemoryBank}~\citep{zhong2024memorybank}.
We preserve the original MemoryBank design, including user-specific long-term
memory banks, dated dialogue memories, event- and personality-level summaries,
and FAISS indexing with all-MiniLM-L6-v2 embeddings at top-8 retrieval. This
retains its native memory granularity rather than the unified retrieval
granularity used for other baselines.

\item \textbf{MemSkill}~\citep{zhang2026memskill}.
We use the controller checkpoint released by the original authors, trained on
LoCoMo, and evaluate it zero-shot on our test split without retraining. Qwen3-32B
serves as the executor, with the top-50 memories retrieved for answer generation.

\item \textbf{MemT}~\citep{yue2026memt}.
We follow the official inference setup using the released Mem-T-4B checkpoint,
the default retrieval top-$k$, and the original tool-calling and maximum
tool-step settings. For LongMemEval we build the memory bank with the official
pipeline; for LoCoMo we use the released prebuilt memory bank.

\item \textbf{NanoMemory}~\citep{wu2026back}.
We use its query-time memory construction pipeline with session-level retrieval,
and increase the retrieval depth from the original top-3 to top-10 for a stronger
comparison.

\end{itemize}

\subsection{LLM Judge}
\label{app:llm_judge}
For both \textbf{LongMemEval}~\citep{wu2025longmemeval} and
\textbf{LoCoMo}~\citep{maharana2024evaluating}, we adopt the official
evaluation scripts and judge prompts released with the original papers,
changing only the judge model. Concretely, the judge is
DeepSeek-V4-Pro (OpenAI-compatible API at
\url{https://api.deepseek.com/v1},
accessed in July 2026, temperature 0), which receives the question, reference
answer, and model prediction, and outputs a binary correctness label. We do
not
modify the prompts, decision rules, or answer-matching logic of these scripts;
we only substitute the underlying model. 

Within each benchmark, the judge is applied uniformly to all systems and never
observes which system produced a prediction. Applying the same judge avoids
method-specific evaluation procedures, although model-dependent judge errors
may still occur; these are quantified in Appendix~\ref{app:reliability}.

\section{Reward Details}
\label{app:reward_details}

This appendix supplements the reward design in Section~\ref{sec:training} with
the action-reward analysis (Section~\ref{app:action_reward}), the
quality-reward rubric and judge configuration
(Section~\ref{app:quality_rubric}), and the reward-mixture and curriculum settings
(Section~\ref{app:reward_hyper}).

\subsection{Action Reward Analysis}
\label{app:action_reward}

We use $\eta=0.915$ in all experiments. This fixed asymmetric weight gives
greater priority to preserving gold evidence while retaining a positive
incentive to remove non-gold context. We treat it as a recall-prioritizing
preference rather than a calibrated estimate of downstream error costs.

\paragraph{Observed window composition.}
The raw RL pool is strongly imbalanced across windows: 91.02\% contain no gold
message. We address this prompt-level imbalance by sampling equal numbers of
gold-bearing and non-gold-only windows. Imbalance also remains within mixed
windows, where the median and 95th-percentile non-gold/gold ratios are 5 and 7.
Category-wise aggregation removes this within-window prevalence effect. All
observed mixed windows contain at most eight gold messages; consequently, under
our setting, correcting one dropped gold message changes $R_{\mathrm{act}}$
more than the maximum possible change contributed by the entire non-gold
category. This action-level separation holds for every mixed window in the raw
RL pool, balanced RL set, and validation set.

\paragraph{Reward-ordering diagnostic.}
We next re-score the same sampled behaviors under alternative action rewards.
The SFT reference policy produces eight rollouts for each of 218 validation
prompts, yielding 1,744 outputs, of which 1,675 (96.04\%) pass the training-time
format/schema gate. The gate checks only structural validity and does not judge
action correctness. Since pairs whose two category accuracies move in the same
direction do not identify their relative weighting, we evaluate format-valid
conflict pairs in which one rollout retains more gold but rejects less non-gold
context.

For each rollout, let $A_{\mathrm{g}}$ denote the fraction of gold messages
assigned \textsc{Keep}, and $A_{\mathrm{n}}$ the fraction of non-gold messages
assigned \textsc{Drop}. Let $N_{\mathrm{g}}$ and $N_{\mathrm{n}}$ denote the
respective numbers of messages in the window. We compare two diagnostic
baselines against the action reward used for training
(Eq.~\ref{eq:action_reward}):
\begin{align}
R_{\mathrm{Micro}}
&= \frac{N_{\mathrm{g}} A_{\mathrm{g}} +
N_{\mathrm{n}} A_{\mathrm{n}}}{N_{\mathrm{g}} + N_{\mathrm{n}}},
\label{eq:micro_action_reward} \\
R_{\mathrm{Symmetric}}
&= 0.5 A_{\mathrm{g}} + 0.5 A_{\mathrm{n}},
\label{eq:symmetric_action_reward} \\
R_{\mathrm{Ours}} = R_{\mathrm{act}}
&= \eta A_{\mathrm{g}} + (1-\eta) A_{\mathrm{n}},
\qquad \eta=0.915.
\label{eq:ours_action_reward}
\end{align}
The micro reward is ordinary message-level action accuracy, so each category's
contribution is proportional to its message count. The symmetric reward gives
equal weight to gold retention and non-gold rejection, whereas our asymmetric
reward prioritizes gold retention. By construction, each conflict pair contains
one rollout with higher $A_{\mathrm{g}}$ but lower $A_{\mathrm{n}}$ than the
other. Let $o_{\mathrm{g}}$ denote this rollout and $o_{\mathrm{n}}$ the other.
An
\emph{inversion} occurs when
$R(o_{\mathrm{g}}) < R(o_{\mathrm{n}})$: the reward favors better non-gold
rejection over better gold retention. A \emph{tie} occurs when
$R(o_{\mathrm{g}}) = R(o_{\mathrm{n}})$, so the reward provides no preference
between the two. Table~\ref{tab:action_reward_ordering} reports the fractions
of conflict pairs exhibiting each outcome.

\begin{table}[!t]
\centering
\caption{Reward ordering on fixed, format-valid SFT rollouts. Inversion and tie
rates are computed over gold/non-gold conflict pairs (lower is better). Gold
gap is the mean reward difference between the best gold-retaining rollout and
the best rollout that drops gold (higher is better).}
\label{tab:action_reward_ordering}
\small
\setlength{\tabcolsep}{5pt}
\begin{tabular}{lccc}
\toprule
Reward & Inversion & Tie & Gold gap \\
\midrule
Micro     & 54.2\% & 45.8\% & 0.173 \\
Symmetric & 0.0\%  & 16.9\% & 0.493 \\
Ours      & 0.0\%  & 0.0\%  & 0.901 \\
\bottomrule
\end{tabular}
\end{table}

Micro averaging either reverses or ties every observed conflict pair.
Symmetric category balancing removes the reversals but leaves 16.9\% as ties,
whereas our asymmetric reward removes both and substantially enlarges the gold
gap. This diagnostic supports a recall-prioritizing regime rather than uniquely
identifying the submitted numerical value.

\paragraph{Downstream perturbation.}
Finally, we test whether this preference agrees with downstream answer
sensitivity. On 40 validation questions, we compare a full-gold memory with
paired variants that remove gold evidence or add non-gold messages, using the
same Qwen3-32B answer model and correctness judge as in the main evaluation.
Multiple deletions and random additions are first averaged within each question;
confidence intervals are then computed by question-level paired bootstrap.
Table~\ref{tab:action_reward_perturbation} summarizes the resulting changes in
answer accuracy and input length.

\begin{table}[!t]
\centering
\caption{Controlled memory perturbations on the validation set. $\Delta$ is
the paired accuracy change from full gold, C$\rightarrow$I is the fraction of
initially correct answers flipped to incorrect, and Tok. is the mean input
length computed with the Qwen3-32B tokenizer.}
\label{tab:action_reward_perturbation}
\small
\setlength{\tabcolsep}{2.5pt}
\begin{tabular}{lrrrr}
\toprule
Memory & Acc. (\%) & $\Delta$ (pp) & C$\rightarrow$I (\%) & Tok. \\
\midrule
Full gold       & 77.5 & ---   & ---  & 728 \\
Drop one gold   & 18.0 & -59.5 & 80.6 & 603 \\
Drop all gold   & 10.0 & -67.5 & 93.5 & 122 \\
$+$1 non-gold   & 85.8 &  +8.3 &  3.2 & 956 \\
$+$5 non-gold   & 79.2 &  +1.7 &  6.5 & 2,021 \\
$+$all non-gold & 42.5 & -35.0 & 48.4 & 30,385 \\
\bottomrule
\end{tabular}
\end{table}

Removing one gold message reduces accuracy by 59.5 percentage points
(paired 95\% CI: $[-74.2,-42.8]$) and flips 80.6\% of initially correct
answers. Retaining one or five additional non-gold messages produces no
statistically detectable degradation, whereas retaining all local non-gold
context reduces accuracy by 35.0 points (95\% CI: $[-52.5,-20.0]$) while
increasing the mean input to 30,385 tokens. The same qualitative pattern holds
for single-gold and multi-gold questions. These interventions support a strong
but non-exclusive preference for gold retention: evidence omission is highly
destructive, limited over-retention is comparatively tolerable, and extreme
over-retention remains harmful.

Together, these diagnostics validate the direction and operating regime of the
asymmetric reward rather than claiming that $\eta=0.915$ is a calibrated
optimum. Establishing end-to-end sensitivity to the exact weight would require
retraining policies under alternative settings.

\FloatBarrier

\subsection{Quality Reward: Scoring Rubric and Judge Configuration}
\label{app:quality_rubric}

\paragraph{Judge model.}
We compute the quality reward $R_{\mathrm{qual}}$ (Eq.~\ref{eq:quality_reward})
with Qwen3-32B~\citep{yang2025qwen3technicalreport} at temperature~0 and a
maximum of 1,024 output tokens. Thinking is enabled with a 512-token budget;
each request has a 180\,s timeout and up to three retries. During training, the
judge receives the source message, query, gold answer, and referenced reasoning
chain. The gold answer and reasoning chain are used only to assess utility when
computing the training reward and are unavailable at inference time. The full
judge prompt is given in Listing~\ref{lst:reward-judge-prompt}.

The faithfulness score $f_x$ grades whether the compressed content is grounded
in the source message, while the utility score $u_x$ grades how much it helps
answer query~$q$. Table~\ref{tab:quality_reward_rubric} gives the complete
three-level rubric for both dimensions.

\begin{table}[H]
\centering
\caption{Scoring rubric for the faithfulness and utility components of the
quality reward.}
\label{tab:quality_reward_rubric}
\footnotesize
\setlength{\tabcolsep}{3pt}
\begin{tabularx}{\columnwidth}{@{}lcX@{}}
\toprule
Dimension & Score & Criterion \\
\midrule
Faithfulness
& 0 & Contains unsupported, fabricated, contradictory, or over-inferred
information not grounded in the source message. \\[2pt]
& 1 & Mostly faithful but contains minor ambiguity, imprecise wording, or
mild over-generalization not explicitly supported by the source. \\[2pt]
& 2 & Fully faithful; all content is explicitly supported by the source with
no contradiction, hallucination, or over-inference. \\
\midrule
Utility
& 0 & Irrelevant or misleading; not helpful for the reference answer or
reasoning chain. \\[2pt]
& 1 & Somewhat helpful; provides indirect support, useful background,
disambiguating context, or partially relevant evidence. \\[2pt]
& 2 & Strongly helpful; preserves key evidence, constraints, or context
needed to reach, verify, or explain the answer. \\
\bottomrule
\end{tabularx}
\end{table}

\paragraph{Human validation.}
\label{app:reward_judge_validation}

We randomly sampled 100 message-level compression instances from rollouts
generated during training. Three expert annotators independently scored each
instance for faithfulness and utility using the same scoring rubrics as the
reward judge, while remaining blind to the judge's predictions. The final
human label was determined by majority vote, with a fourth expert resolving
any three-way tie. Table~\ref{tab:reward_judge_agreement} reports agreement
between the reward judge and the resulting human labels.

\begin{table}[H]
\centering
\caption{Exact-match agreement between the reward judge and
expert-aggregated human labels on 100 randomly sampled training rollouts.}
\label{tab:reward_judge_agreement}
\small
\begin{tabular}{lc}
\toprule
Dimension & Agreement \\
\midrule
Faithfulness & 91\% (91/100) \\
Utility      & 85\% (85/100) \\
\bottomrule
\end{tabular}
\end{table}

\FloatBarrier

\subsection{Reward-Mixture and Curriculum Hyperparameters}
\label{app:reward_hyper}

Table~\ref{tab:reward_hyper} lists the reward-mixture and curriculum settings
used in training.
The action-reward weight encodes a relative preference for gold retention
rather than a calibrated downstream cost.
\begin{table}[H]
\centering
\caption{Reward-mixture and curriculum hyperparameters.}
\label{tab:reward_hyper}
\small
\begin{tabular}{llc}
\toprule
Component & Parameter & Value \\
\midrule
Action reward
& Gold-retention weight $\eta$ & 0.915 \\
\midrule 
Reward mixture
& Phase~1 quality weight $\lambda_{\mathrm{qual}}$ & 0.0 \\
& Phase~2 quality weight $\lambda_{\mathrm{qual}}$ & 0.5 \\
\midrule
Curriculum
& Format-validity threshold $\tau$  & 0.99 \\
\bottomrule
\end{tabular}
\end{table}

\FloatBarrier

\section{Error Attribution and Evaluation Reliability}
\label{app:case-study}

This appendix supplements Section~\ref{sec:analysis} with representative
failure examples and an assessment of evaluation reliability.

\begin{table}[t]
\centering
\small
\caption{Representative failure examples across pipeline stages and benchmarks.}
\label{tab:app_cases}
\setlength{\tabcolsep}{3pt}
\renewcommand{\arraystretch}{1.15}
\begin{tabularx}{\textwidth}{@{}clllX@{}}
\toprule
\textbf{ID} & \textbf{Bench.} & \textbf{Type} & \textbf{Stage} & \textbf{Phenomenon} \\
\midrule
\multicolumn{5}{c}{\textit{Retrieval Miss}} \\
\midrule
R1 & LoCoMo & Single-hop & Retrieval & Gold evidence (phone navigation malfunction) absent from all retrieved candidate windows; model substitutes a semantically similar self-checkout malfunction. \\
R2 & LoCoMo & Multi-hop & Retrieval & Two gold items needed (mansion + Ferrari); only Ferrari recalled. Model reports what it sees but cannot complete the set. \\
\midrule
\multicolumn{5}{c}{\textit{Editing Loss}} \\
\midrule
C1 & LoCoMo & Temporal & Editing & ``last week'' rewritten to message date (June 6); gold answer is the week before. Temporal anchor destroyed by compression. \\
C2 & LoCoMo & Temporal & Editing & ``Yesterday'' rewritten to message date (Oct 4) instead of Oct 3. One-day offset from conflating report time with event time. \\
C3 & LongMemEval & Multi-session & Editing & Pre-approval amount (\$350k) retained but sale price (\$325k) dropped; QA cannot compute the \$25k difference. \\
C4 & LongMemEval & Multi-session & Editing & HelloFresh 40\% discount retained but UberEats 20\% dropped; comparison unanswerable with one operand. \\
\midrule
\multicolumn{5}{c}{\textit{QA Reasoning Error}} \\
\midrule
Q1 & LoCoMo & Temporal & QA & Memory preserves ``a few days ago'' intact, but QA outputs the message date instead of computing the offset. \\
Q2 & LongMemEval & Multi-session & QA & Both operands present (Alex 21, user 32); model outputs ``cannot determine'' instead of computing $32-21=11$. \\
\bottomrule
\end{tabularx}
\end{table}

\subsection{Representative Failure Examples}
\label{app:cases}

Table~\ref{tab:app_cases} presents representative examples for each failure type.
Cases prefixed R denote retrieval miss, C denotes editing loss, and Q
denotes QA reasoning error.

The examples illustrate distinct information bottlenecks. Retrieval misses
remove required facts before construction, editing losses either discard an
operand or corrupt a temporal anchor, and QA errors occur despite sufficient
evidence in the constructed memory.

\subsection{Human Evaluation and Judge Noise Analysis}
\label{app:reliability}

To assess the reliability of automatic evaluation, three expert annotators
independently audit every LLM judge label on both benchmarks, with final
decisions determined by majority vote. For each label, annotators inspect the
question, reference answer, model prediction, and automatic decision. They
also consult the complete interaction history and benchmark-provided gold
evidence when resolving correctness or a suspected reference-answer error.
Entity, numerical, and cross-method consistency signals provide additional
checks for judge-positive samples rather than replacing manual review. On LoCoMo, we
additionally identify 19 questions with gold-answer annotation errors that
are excluded only from the stage-level error-attribution analysis. They remain
in the 314-question raw and judge-corrected LJ calculations reported below.

A \emph{false negative} (FN) is a correct prediction judged incorrect; a
\emph{false positive} (FP) is an incorrect prediction judged correct.
Table~\ref{tab:judge_noise} reports the results for all 13 methods appearing
in the main evaluation (12 on LoCoMo, as Oracle Session is not defined there).
We compute
$\text{Corrected LJ}=\text{Raw LJ}+(\mathrm{FN}-\mathrm{FP})/N$ from the
underlying binary labels and then round the raw and corrected scores
independently to two decimals. The 19 LoCoMo questions with gold-answer errors
remain in both columns, so the denominator is 314; they are excluded only from
the subsequent stage-level error attribution.

\begin{table}[ht]
\centering
\small
\caption{Judge-noise audit across all methods in the main evaluation. FN and FP
are false-negative and false-positive counts under the human majority decision;
Corr.\ LJ applies the corresponding label corrections.}
\label{tab:judge_noise}
\setlength{\tabcolsep}{3pt}
\renewcommand{\arraystretch}{1.05}
\resizebox{\linewidth}{!}{%
\begin{tabular}{@{}l|cccc|cccc@{}}
\toprule
& \multicolumn{4}{c|}{\textbf{LongMemEval} (100 questions)}
& \multicolumn{4}{c}{\textbf{LoCoMo} (314 questions)} \\
\cmidrule(lr){2-5} \cmidrule(lr){6-9}
\textbf{Method}
& Raw LJ & FN & FP & Corr.\ LJ
& Raw LJ & FN & FP & Corr.\ LJ \\
\midrule
\multicolumn{9}{c}{\textit{Gold Context}} \\
\midrule
Oracle Turn      & 0.82 & 3 & 0 & 0.85 & 0.75 & 24 & 0 & 0.82 \\
Oracle Session   & 0.84 & 0 & 1 & 0.83 & -- & -- & -- & -- \\
\midrule
\multicolumn{9}{c}{\textit{Basic Retrieval}} \\
\midrule
RAG Top20        & 0.71 & 2 & 2 & 0.71 & 0.67 & 20 & 0 & 0.74 \\
RAG Top50        & 0.75 & 2 & 2 & 0.75 & 0.66 & 25 & 0 & 0.74 \\
\midrule
\multicolumn{9}{c}{\textit{Write-Time Construction}} \\
\midrule
LightMem         & 0.79 & 2 & 1 & 0.80 & 0.47 & 36 & 1 & 0.58 \\
StructMem        & 0.82 & 2 & 1 & 0.83 & 0.52 & 29 & 0 & 0.61 \\
Mem0             & 0.64 & 1 & 3 & 0.62 & 0.45 & 35 & 0 & 0.56 \\
MemoryBank       & 0.68 & 0 & 1 & 0.67 & 0.41 & 25 & 0 & 0.49 \\
MemSkill         & 0.57 & 1 & 2 & 0.56 & 0.43 & 23 & 0 & 0.50 \\
MemT             & 0.61 & 5 & 1 & 0.65 & 0.57 & 23 & 0 & 0.64 \\
\midrule
\multicolumn{9}{c}{\textit{Query-Time Construction}} \\
\midrule
NanoMemory       & 0.80 & 0 & 2 & 0.78 & 0.68 & 13 & 0 & 0.72 \\
\textbf{LazyMem} (32B)  & 0.93 & 0 & 0 & 0.93 & 0.69 & 18 & 0 & 0.75 \\
\textbf{LazyMem}-4B     & 0.85 & 0 & 0 & 0.85 & 0.68 & 22 & 0 & 0.75 \\
\bottomrule
\end{tabular}%
}
\end{table}

\paragraph{LongMemEval.}
Across 13 methods $\times$ 100 questions = 1,300 labels, we identify 18 FN and
16 FP (overall noise rate 2.62\%). False negatives and false positives are
roughly balanced, so per-method corrections range from $-2$ to $+4$ percentage
points. Both \textsc{LazyMem} variants have zero judge noise; their reported
scores are exact. The two variants remain the top two non-oracle methods after
correction, so the central comparison is unchanged.

\paragraph{LoCoMo.}
Across 12 methods $\times$ 314 questions = 3,768 labels, judge noise is higher
and strongly directional: of 294 flipped labels, 293 are false negatives and
only 1 is a false positive (overall noise rate 7.80\%). The raw judge therefore
systematically underestimates all methods. The bias is consistent across methods
(FN rates range from 4.14\% to 11.46\% per method, with net corrections of
$+4.14$ to $+11.15$ percentage points); although some pairwise rankings change
after correction, both \textsc{LazyMem} variants remain tied for the best
corrected non-oracle LJ. Additionally, we identify 19 questions where the gold
answer itself contains annotation errors (wrong subject, incorrect date, or
unsupported temporal claims). We retain these questions in the main and
judge-corrected LJ scores, but exclude them uniformly when attributing confirmed
errors to retrieval, editing, or QA stages.

\paragraph{Judge--human agreement.}
Treating the human majority-vote decision as the reference, the automatic judge
agrees on 1,266 of 1,300 LongMemEval labels (97.38\%) and 3,474 of 3,768 LoCoMo
labels (92.20\%), for 4,740 of 5,068 labels overall (93.53\%). These values
measure agreement between the automatic judge and the human majority decision.

\paragraph{Sources of judge noise.}
The dominant noise patterns, shared across both benchmarks, are:
\begin{itemize}
\item Semantically equivalent answers rejected due to surface-form differences
(e.g., ``3'' vs.\ ``3 weddings'', ``45 minutes'' vs.\ ``45 minutes each way'').
\item Keyword overlap masking factual errors, leading to false positives
(e.g., correct entity mentioned alongside an incorrect temporal quantity).
\item Relative temporal expressions (``last week'', ``tomorrow'') not recognized
as resolving to the gold date.
\item Cautious qualifications (``no exact date is given'') causing the judge to
ignore a preceding correct answer.
\end{itemize}

\paragraph{Impact on reported results.}
The main table retains the prespecified automatic evaluation metric, while the
human-corrected results serve as a sensitivity analysis. The audit confirms
that (1) the two \textsc{LazyMem} variants
remain the top two non-oracle methods on both benchmarks after correction,
(2) judge noise does not inflate \textsc{LazyMem}'s reported performance
(zero label errors on LongMemEval and only false negatives on LoCoMo), and
(3) the error attribution analysis in
Section~\ref{sec:analysis} excludes human-confirmed judge noise and the 19
gold-answer annotation errors before classifying failures.

\section{Method-Specific Prompts and Output Schemas}
\label{app:prompts}

This section gives the complete text of the prompts introduced by our method
for window annotation, training-time quality scoring, and answer generation.
The final-answer correctness evaluation instead uses the unchanged official
benchmark prompts described in Section~\ref{app:llm_judge}.
Braced expressions such as \texttt{\{query\}} are runtime placeholders. The
headings in the source listings (e.g., ``SYSTEM PROMPT'') separate messages for
presentation and are not themselves sent to the model.

\subsection{Window Annotation Prompt}
\label{app:window-annotation-prompt}

Listing~\ref{lst:window-annotation-prompt} is used to annotate a local
chronological window with query-conditioned \textsc{Keep}/\textsc{Drop}
decisions. The output is required to be a JSON array with one decision for
each non-empty role message in chronological order.

\begin{lstlisting}[
  style=prompt,
  caption={System message and user-message template for local window annotation.},
  label={lst:window-annotation-prompt}
]
SYSTEM PROMPT

You are a local memory editing model.

You will receive a user query and one local chronological memory window.
Your task is to output a sequence of local editing decisions for this local window only, following the original message order.

Return exactly one valid JSON array.
Each item in the array must correspond to one non-empty role message in previous_bridge, core_messages, and next_bridge.
Each item in the array must contain exactly three fields: op, compressed_content, and reason.

The only allowed decision operations are KEEP and DROP.
For KEEP, retain dialogue content that is relevant to the query and useful for answering the query.
KEEP may be realized as verbatim keeping, key-span extraction, or concise compression depending on each message.
For DROP, compressed_content must be an empty string.

Use only the provided window content.
Do not output markdown, explanations, or any extra text.
Your output must be directly parseable by json.loads().


USER PROMPT TEMPLATE

# Task
Given the following query-conditioned memory window, produce a sequence of local editing decisions.

# Required Output
Return only a JSON array with the following structure:
[
  {
    "op": "KEEP | DROP",
    "compressed_content": "empty string if DROP; otherwise retained content in one KEEP style: verbatim keeping, key-span extraction, or concise compression",
    "reason": "brief local reason"
  }
]

# Local Editing Rules
1. The output array must contain one decision for every non-empty role message in previous_bridge, core_messages, and next_bridge.
2. The decisions must follow the original chronological order: previous_bridge, then core_messages, then next_bridge.
3. Each decision must contain exactly three fields: op, compressed_content, and reason.
4. op must be either KEEP or DROP.
5. KEEP means preserving dialogue content that is relevant to the query and useful for the main model to answer the question.
6. DROP means the message is locally irrelevant and should not enter the final memory.
7. For KEEP, choose the most appropriate retention style for each message:
   - verbatim keeping: keep the original message text when it is short and directly useful;
   - key-span extraction: keep only the query-relevant span when only part of the message is relevant;
   - concise compression: faithfully rewrite the message into a shorter form when it is useful but too long, or only partially relevant.
8. For any KEEP style, remain faithful to the source meaning and do not introduce unsupported facts.
9. If the whole window is irrelevant, still output one decision for every non-empty role message, and set every op to DROP.
10. When op is DROP, compressed_content must be an empty string.
11. If a message may contain answer evidence, user facts, temporal updates, or content related to the query entity/attribute, prefer KEEP even when relevance is uncertain.
12. Return only the JSON array. Do not output markdown, comments, or extra text.
13. The output must be directly parseable by json.loads().

# Query
Query Time: {query_time}
Query: {query}

# History Window
{window}
\end{lstlisting}

\subsection{Memory-quality Reward Judge Prompt}
\label{app:reward-judge-prompt}

Listing~\ref{lst:reward-judge-prompt} is used by $R_{\mathrm{qual}}$ to
score a kept constructed memory independently for faithfulness and utility.

\begin{lstlisting}[
  style=prompt,
  caption={System message, scoring rubric, and output schema for the memory-quality judge.},
  label={lst:reward-judge-prompt}
]
SYSTEM PROMPT

You are a strict reward judge for memory construction. Evaluate whether a kept
constructed_memory is faithful and useful. Keep the two dimensions strictly
separate: faithfulness only judges the relationship between constructed_memory
and source_messages; utility judges how helpful constructed_memory is for the
gold_answer and referenced_reasoning_chain. Return exactly one valid JSON object.
Do not output Markdown, explanations, or extra text.


USER PROMPT TEMPLATE

{
  "query_time": "{query_time}",
  "query": "{query}",
  "gold_answer": "{gold_answer}",
  "referenced_reasoning_chain": "{referenced_reasoning_chain}",
  "source_messages": [
    {
      "role": "{role}",
      "content": "{source_message_content}",
      "timestamp": "{timestamp}"
    }
  ],
  "constructed_memory": "{compressed_memory}",
  "evaluation_rules": [
    "Score faithfulness using only source_messages. Do not use gold_answer or referenced_reasoning_chain for faithfulness.",
    "Each source message may include a timestamp. Treat that timestamp as part of the source evidence when checking dates, chronology, ordering, and time intervals.",
    "Treat query_time as the time when the query was asked. Compare it with source message timestamps when judging temporal relevance, recency, chronology, and knowledge updates.",
    "Faithfulness checks whether constructed_memory contains unsupported, fabricated, contradictory, or over-inferred information relative to source_messages, including their timestamps.",
    "Score utility using query, gold_answer, and referenced_reasoning_chain together.",
    "Utility checks whether constructed_memory helps answer the query, reach or verify gold_answer, or perform the reasoning in referenced_reasoning_chain.",
    "Utility does not require a direct one-hop contribution. Indirectly useful evidence, constraints, background, disambiguation, chronology, entities, or context can receive utility credit if they help the final answer or reasoning chain.",
    "Score faithfulness and utility independently; a faithful but useless memory may have faithfulness=2 and utility=0.",
    "A useful but unfaithful memory must have faithfulness=0; do not raise faithfulness because it matches gold_answer or referenced_reasoning_chain.",
    "If constructed_memory is only generic background or repeats information irrelevant to gold_answer and referenced_reasoning_chain, assign a low utility score.",
    "If constructed_memory preserves key direct or indirect evidence needed by gold_answer or referenced_reasoning_chain, assign a high utility score.",
    "When uncertain, choose the lower score."
  ],
  "rubric": {
    "faithfulness": {
      "0": "Contains information unsupported by source_messages, fabrication, clear contradiction, or states uncertain information as fact.",
      "1": "Mostly faithful, but has minor ambiguity, mild over-generalization, imprecise wording, or details not explicitly supported by source_messages.",
      "2": "Fully faithful; all key information is explicitly supported by source_messages, with no contradiction, hallucination, or over-inference."
    },
    "utility": {
      "0": "Not helpful for gold_answer or referenced_reasoning_chain, irrelevant, or misleading for the final answer or reasoning.",
      "1": "Somewhat helpful for gold_answer or referenced_reasoning_chain, including indirect support, useful background, disambiguating context, chronology, entities, constraints, or partially relevant evidence.",
      "2": "Strongly helpful for gold_answer or referenced_reasoning_chain; preserves key direct or indirect evidence, constraints, facts, or context needed to reach, verify, or explain the answer."
    }
  },
  "required_output": {
    "faithfulness": 0,
    "utility": 0,
    "unsupported_claims": [],
    "useful_evidence": [],
    "reason": "Brief reason in English"
  }
}
\end{lstlisting}

\subsection{Answer Generation Prompt}
\label{app:answer-generation-prompt}

Listing~\ref{lst:answer-generation-prompt} gives the shared answer-generation
prompt. The compressed and gold/text memory conditions differ only in
the description and rendering of the memory block.

\balance
\begin{lstlisting}[
  style=prompt,
  caption={System message and user-message templates for answer generation.},
  label={lst:answer-generation-prompt}
]
SYSTEM PROMPT

Answer the user question concisely and faithfully from the provided memory.


USER PROMPT TEMPLATE FOR COMPRESSED MEMORY

You are answering a user question using compressed chronological conversation memory.
Use the memory when it is relevant. If the memory is insufficient, say what is missing instead of inventing facts.

# Query
Query Time: {query_time}
Query: {query}

# Memory
[{timestamp}] {role}: {compressed_content}
...

# Answer


USER PROMPT TEMPLATE FOR GOLD/TEXT MEMORY

You are answering a user question using chronological conversation memory.
Use the memory when it is relevant. If the memory is insufficient, say what is missing instead of inventing facts.

# Query
Query Time: {query_time}
Query: {query}

# Memory
{memory_text}

# Answer
\end{lstlisting}

\end{document}